\documentclass{article}
\usepackage{iclr2027_conference,times}
\usepackage[T1]{fontenc}
\usepackage[utf8]{inputenc}
\usepackage{amsmath,amssymb,amsthm,bm}
\usepackage{graphicx,booktabs,array}
\usepackage{algorithm,algorithmic}
\usepackage{placeins}
\usepackage{microtype}
\usepackage[hidelinks]{hyperref}
\usepackage{url}
\hypersetup{pdftitle={Stay on the Manifold: Hierarchical Bayesian Trajectory-Family Transfer for Sparse-Target Reconstruction and Sampling},pdfauthor={Elvis Han Cui; Yuhan Pan; Xin-Ya Zhang; Bingbin Li; Yanan Li; Donghui Wang; Weng Kee Wong},pdfsubject={Preprint},pdfkeywords={trajectory transfer, empirical Bayes, few-shot sampling}}
\newcommand{\mat}[1]{\mathbf{#1}}
\newcommand{\vect}[1]{\boldsymbol{#1}}
\newcommand{\R}{\mathbb{R}}
\newcommand{\E}{\mathbb{E}}
\newcommand{\phiv}{\vect{\phi}}
\DeclareMathOperator{\tr}{tr}
\DeclareMathOperator{\diag}{diag}
\DeclareMathOperator{\dist}{dist}
\DeclareMathOperator*{\argmin}{arg\,min}
\newtheorem{proposition}{Proposition}

\title{Stay on the Manifold: Hierarchical Bayesian Trajectory-Family Transfer for Sparse-Target Reconstruction and Sampling}
\iclrfinalcopy
\renewcommand{\headrulewidth}{0pt}
\date{November 6, 2026}
\author{Elvis Han Cui, Yuhan Pan, Xin-Ya Zhang, Bingbin Li,
Yanan Li, Donghui Wang, Weng Kee Wong}
\makeatletter
\renewcommand{\maketitle}{%
  \begingroup
  \centering
  {\LARGE\bfseries\@title\par}
  \vspace{14pt}
  {\normalsize
    Elvis Han Cui\textsuperscript{1,*}\quad
    Yuhan Pan\textsuperscript{2}\quad
    Xin-Ya Zhang\textsuperscript{3}\quad
    Bingbin Li\textsuperscript{4}\\[4pt]
    Yanan Li\textsuperscript{5}\quad
    Donghui Wang\textsuperscript{4}\quad
    Weng Kee Wong\textsuperscript{1}\par}
  \vspace{9pt}
  {\small
    \textsuperscript{1}Department of Biostatistics, Fielding School of Public Health,\\
    University of California, Los Angeles, Los Angeles, CA, USA\\[3pt]
    \textsuperscript{2}Key Laboratory of Intelligent Perception and Human-Machine Collaboration,\\
    ShanghaiTech University, Shanghai, China\\[3pt]
    \textsuperscript{3}Center for Interdisciplinary Studies, Westlake University, Hangzhou, China\\[3pt]
    \textsuperscript{4}Department of Computer Science, Zhejiang University, Hangzhou, China\\[3pt]
    \textsuperscript{5}Research Center for Frontier Fundamental Studies,\\
    Zhejiang Lab, Hangzhou, China\par}
  \vspace{7pt}
  {\small\textsuperscript{*}Corresponding author: \href{mailto:elviscuihan@g.ucla.edu}{\texttt{elviscuihan@g.ucla.edu}}\par}
  \vspace{7pt}
  {\small\@date\par}
  \endgroup
  \vspace{10pt}
}
\makeatother

\begin{document}
\maketitle
\begin{abstract}
Where should reconstructions and new samples lie when a few observations reveal only fragments of a trajectory? We propose hierarchical Bayesian trajectory principal-manifold sampling (HB-TPM), which learns a family of smooth curves from ordered source tasks and adapts it to sparse targets. For unordered vectors, alternating progression assignments and coefficient updates estimate trajectory structure between and beyond the observed shots under the learned source family. Observation-aware extensions connect the trajectory prior to photon arrivals and image-encoded cardiac cycles. Our analysis characterizes the coefficient directions supplied by the prior and bounds sampling error conditional on curve recovery. Across 30 held-out vector tasks, HB-TPM reduces mean sliced-Wasserstein distance by 21.0--41.8\% relative to a source-adapted VAE. Prior-strength and sampling ablations identify coefficient shrinkage as the main supported mechanism. On 100 simulated photon records, adaptation lowers mean relative trajectory RMSE by 50.2\% versus target-only fitting and 23.6\% versus source-mean prediction. Across three training seeds and five folds on 98 cardiac sequences, HB-TPM with latent-MSE training achieves the best mean reconstruction among the tested configurations. Together, the results demonstrate trajectory-family transfer for sparse-target reconstruction and trajectory-centered sampling when source and target tasks share comparable progression structure.
\end{abstract}
\section{Introduction}
Sparse-target reconstruction and sampling face a structural question: where should missing states and additional samples lie when the observed shots cover only fragments of the relevant variation? Metric-based methods transfer representations or comparison rules \citep{vinyals2016matching,snell2017prototypical}, while interpolation, feature synthesis, and generative adaptation construct new examples \citep{chawla2002smote,zhang2018mixup,hariharan2017low,ojha2021few}. Linear interpolation stays within the observed convex hull, and ambient perturbations can spread away from curved high-density regions. Recovering missing structure is therefore central to estimating a useful trajectory and sampling distribution.

Learned representations organize high-dimensional observations in simpler geometries \citep{bengio2013representation,pope2021intrinsic}. We study tasks with a smooth progression, such as an experimental time course or a cardiac cycle. Each task follows its own curve within the representation space. Sparse observations leave many possible continuations, even under smoothness constraints. Related, well-observed tasks offer population-level information about which continuations are plausible.

\begin{figure}[t]
\centering
\includegraphics[width=\linewidth]{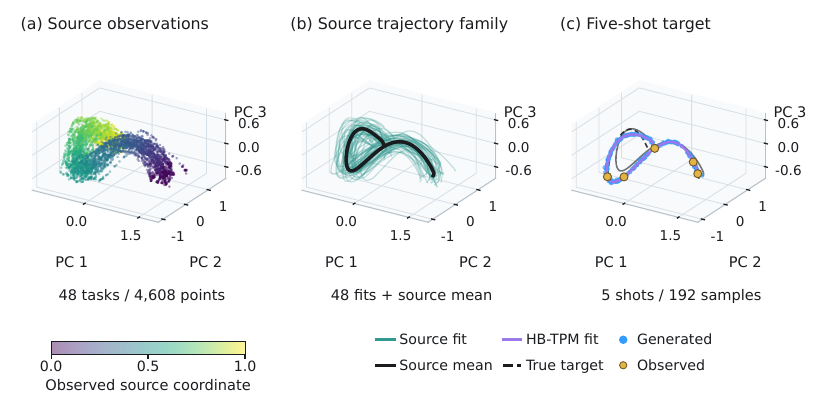}
\caption{Learning a family before adapting a target. (a) All 48 synthetic source tasks, colored by observed progression. (b) Their fitted curves and coefficient-mean curve (black). (c) Five unordered observations (gold), HB-TPM's fitted target (purple) and 192 generated samples (blue), compared with the source mean (solid) and true target (dashed). All panels share a source-only PCA map retaining 89.1\% of source variance. Estimation remains eight-dimensional. This is the first held-out target, not a selected winning case. Source spread and generated points are not credible regions.}
\label{fig:sourcefamily3d}
\end{figure}

Functional-data analysis recovers sparse individual trajectories from population structure \citep{yao2005functional}, and hierarchical meta-learning formalizes transfer through shared priors \citep{grant2018recasting}. We transfer a \emph{family of curves}: sources share a broad shape while retaining task-specific coefficients, and target observations adapt this family to a new trajectory (Figure~\ref{fig:sourcefamily3d}). Comparable source progression coordinates make the between-task structure estimable.

Hierarchical Bayesian trajectory principal-manifold sampling (HB-TPM) implements this transfer with a finite-basis empirical-Bayes model. Sources determine a coefficient mean and covariance. Alternating latent progression assignments and Gaussian coefficient updates adapt the curve to unordered target vectors. The prior stabilizes directions that the target leaves weakly observed, and the fitted curve supports sampling beyond the observed shots.

Observation-aware extensions pair the same transfer principle with an event quasi-likelihood for diffusion inference and a learned image encoder for periodic cardiac reconstruction. These extensions use observed acquisition time or cycle phase and test how the observation model shapes target adaptation.

Our contributions are (i) a trajectory-transfer procedure combining coefficient priors, latent target coordinates, and curve-based sampling, with two observation-aware extensions, (ii) an analysis of prior-determined coefficient directions and sampling stability, and (iii) vector, photon, and real-ultrasound evaluations that distinguish curve recovery, target adaptation, and uncertainty calibration. The vector and photon studies establish recovery gains within transferable families. The three-seed cardiac comparison identifies latent-MSE HB-TPM as the best mean reconstruction configuration among those tested.

\section{Related work}
\paragraph{Geometry, principal curves, and interpolation.}
Principal curves describe self-consistent population structure \citep{hastie1989principal}, and local principal curves follow density geometry \citep{ozertem2011local}. HB-TPM uses finite-basis task-progression curves. Statistical manifold estimation and intrinsic-dimension analysis formalize the broader geometric question \citep{fefferman2016testing,levina2004maximum}. SMOTE, mixup, and manifold mixup interpolate observations or hidden states \citep{chawla2002smote,zhang2018mixup,verma2019manifold}, while decoder geometry and adversarial regularization shape latent paths \citep{arvanitidis2018latent,berthelot2018understanding}. Our focus is the distribution of task curves that guides sparse-target completion.

\paragraph{Functional data and Bayesian transfer.}
PACE estimates sparse functions from population structure \citep{yao2005functional,wang2016functional}. Gaussian processes \citep{rasmussen2006gp} and conditional, latent, attentive, and convolutional neural processes \citep{garnelo2018conditional,garnelo2018neural,kim2019attentive,gordon2020convolutional} model related prediction problems. Transformer and diffusion processes enrich their joint predictions \citep{nguyen2022transformer,dutordoir2023neural,dou2025score,hassan2026efficient}. Hierarchical Bayes also provides a perspective on fast adaptation \citep{finn2017model,grant2018recasting}. HB-TPM combines finite-basis Gaussian conditioning with latent progression assignment and curve-based generation. Its distinctive setting is sparse-target adaptation when vector progression coordinates are unobserved; fixed-coordinate functional reconstruction does not address this assignment problem. Appendix~\ref{app:literature} details these connections and the information-matched comparisons they motivate.

\paragraph{Few-shot generative adaptation.}
Feature hallucination and delta encoders transfer within-class variation \citep{hariharan2017low,schwartz2018delta}, while distribution calibration borrows base-class statistics \citep{yang2021free}. Generative adaptation transfers richer source structure \citep{ojha2021few}. Diffusion, latent diffusion, diffusion autoencoders, and flow matching provide flexible generators \citep{ho2020denoising,rombach2022high,preechakul2022diffusion,lipman2023flow}. Recent few-shot work includes DataDream, HypDAE, geodesic-surface transfer, and unified diffusion distillation/adaptation \citep{kim2024datadream,li2025hypdae,han2026few,bahram2026unidad}. Generative transport time need not be physical task progression, and geometric synthesis alone is not ordered trajectory-family inference. Our source-adapted VAE \citep{kingma2014auto} is a pooled-source vector baseline, not an evaluation against these image generators.

\paragraph{Inference from photon arrivals.}
Photon-level Bayesian inference links Brownian diffusion to fluorescence observations \citep{mchale2004bayesian}. We study transfer of time-varying diffusion coefficients across records, using a finite-state event quasi-likelihood and empirical Bayes across sources. The observation approximation and its numerical settings are specified in Appendix~\ref{app:photon}.

\paragraph{Cardiac temporal reconstruction.}
EchoNet-Dynamic studies video-based cardiac function \citep{ouyang2020video}, CAMUS supports cardiac segmentation \citep{leclerc2019deep}, and TED provides full-cycle annotations \citep{painchaud2022temporal}. We use TED to evaluate hidden-frame mask reconstruction from sparse images across the cardiac cycle. Its dense annotations support comparisons of prior adaptation and observation losses within a shared architectural family.

\section{Trajectory-family transfer}
\label{sec:method}
\begin{figure}[t]
\centering
\includegraphics[width=\linewidth]{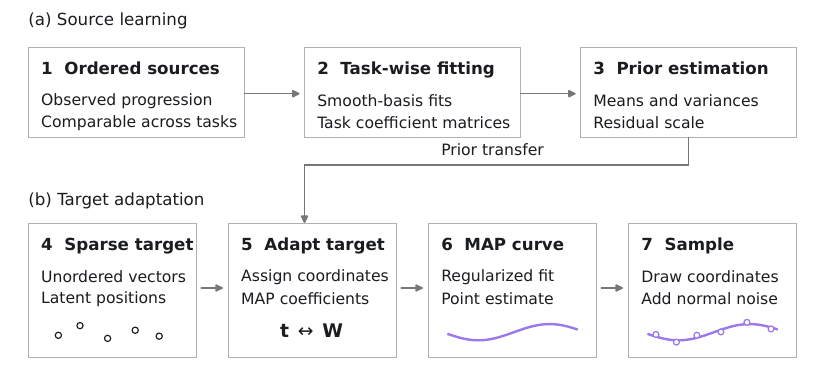}
\caption{Vector HB-TPM workflow. Ordered sources supply a diagonal coefficient prior and residual scale (1--3). Alternating projection assignments and conditional MAP updates fit the target curve (4--6), which is then perturbed for sampling (7). Assignments may coincide. The curve is a point estimate, not a posterior draw. The sketches depict operations rather than experimental results.}
\label{fig:framework}
\end{figure}

HB-TPM represents each task by smooth-basis coefficients. Ordered sources determine which coefficient configurations are plausible, and sparse target observations update that information before sampling. Figure~\ref{fig:framework} summarizes the workflow. We first develop the vector model, then replace its observation model for photon and cardiac data. Algorithm~\ref{alg:hb} in Appendix~\ref{app:implementation} gives the updates. Fixed-representation and non-hierarchical specializations appear in Appendix~\ref{app:imagespecialization}.

\subsection{Observations and working model}
Let source task $j$ provide $\{(t_{ji},\vect{z}_{ji})\}_{i=1}^{n_j}$, where $t_{ji}\in[0,1]$ is an observed progression coordinate and $\vect{z}_{ji}\in\R^D$. A target task provides $\mat{Z}=[\vect{z}_1^\top;\ldots;\vect{z}_K^\top]\in\R^{K\times D}$ without its coordinates. We assume that a common basis $\phiv:[0,1]\to\R^p$ describes the tasks through
\begin{equation}
\vect{\gamma}_{\mat{W}}(t)=\mat{W}^{\top}\phiv(t),\qquad
\mat{W}=[\vect{w}_1,\ldots,\vect{w}_D]\in\R^{p\times D}.
\label{eq:curve}
\end{equation}
Bold upper-case symbols denote matrices and bold lower-case symbols denote vectors. Scalar basis entries are $\phi_b(t)$. The implementation uses the degree-nine Bernstein basis ($p=10$), rather than the simulator's trigonometric basis.

For each ambient coordinate $d$, the working hierarchy is
\begin{equation}
\vect{w}_d\sim\mathcal N(\vect{w}_{0d},\vect{\Lambda}_d),\qquad
\vect{z}_{:d}\mid\vect{w}_d,\vect{t}\sim
\mathcal N(\mat{F}_{\vect{t}}\vect{w}_d,\sigma^2\mat{I}_K),
\quad (\mat{F}_{\vect{t}})_{ib}=\phi_b(t_i).
\label{eq:hierarchy}
\end{equation}
Coordinates are independent under this prior and each $\vect{\Lambda}_d$ is diagonal. Hyperparameters are estimated from sources and held fixed during target inference, giving an empirical-Bayes model. The isotropic likelihood in Eq.~\eqref{eq:hierarchy} is a working likelihood; the benchmark simulator instead generates normal-direction perturbations.

\subsection{Source estimation}
\label{sec:sourcefit}
For each source, form its basis matrix $\mat{F}_j$ and observation matrix $\mat{Z}_j$. Ridge fits yield
\begin{equation}
\widehat{\mat{W}}_j=(\mat{F}_j^\top\mat{F}_j+\lambda_s\mat{I}_p)^{-1}\mat{F}_j^\top\mat{Z}_j.
\label{eq:source}
\end{equation}
The task-average coefficients and their entrywise sample variances define the prior, with a positive variance floor. Figure~\ref{fig:sourcefamily3d}(b) shows the fitted source curves behind these estimates. A clipped source-residual heuristic supplies $\sigma$. Source fitting error is not removed from these plug-in estimates, and the radial scale need not calibrate an isotropic likelihood. Appendix~\ref{app:implementation} specifies the settings.

\subsection{Sparse target adaptation}
With a uniform prior on target grid positions, adaptation minimizes
\begin{equation}
\mathcal J(\mat{W},\vect{t})=
\frac{\|\mat{Z}-\mat{F}_{\vect{t}}\mat{W}\|_F^2}{2\sigma^2}
+\frac12\sum_{d=1}^D(\vect{w}_d-\vect{w}_{0d})^\top
\vect{\Lambda}_d^{-1}(\vect{w}_d-\vect{w}_{0d}).
\label{eq:objective}
\end{equation}
Initialize $\mat{W}$ at the source mean. Each observation is independently assigned to its closest point on a uniform grid $\mathcal G\subset[0,1]$:
\begin{equation}
\widehat t_i\leftarrow\argmin_{t\in\mathcal G}
\|\vect{z}_i-\mat{W}^{\top}\phiv(t)\|^2.
\label{eq:projection}
\end{equation}
Indices may coincide. This is latent coordinate assignment rather than recovery of a permutation. Conditional on these assignments, update each coefficient column by
\begin{align}
\vect{\Sigma}_d&=(\vect{\Lambda}_d^{-1}+\sigma^{-2}\mat{F}_{\widehat{\vect{t}}}^{\top}\mat{F}_{\widehat{\vect{t}}})^{-1},\nonumber\\
\widehat{\vect{w}}_d&=\vect{\Sigma}_d\bigl(\vect{\Lambda}_d^{-1}\vect{w}_{0d}
+\sigma^{-2}\mat{F}_{\widehat{\vect{t}}}^{\top}\vect{z}_{:d}\bigr).
\label{eq:map}
\end{align}
The implementation solves linear systems using 401 grid points and at most 20 alternations, stopping when assignments stabilize. Each block minimizes its part of the objective. The joint optimization remains nonconvex, so the result is a local alternating fit.

\subsection{Sampling from the fitted curve}
Draw $u\sim\operatorname{Unif}[0,1]$ and form the unit tangent
$\vect{v}(u)=\widehat{\vect{\gamma}}'(u)/\|\widehat{\vect{\gamma}}'(u)\|$.
For $D\geq2$ and nonzero tangent and projected direction, let
\begin{equation}
\vect{g}\sim\mathcal N(\vect{0},\mat{I}_D),\quad
\vect{r}(u)=\frac{(\mat{I}_D-\vect{v}(u)\vect{v}(u)^\top)\vect{g}}
{\|(\mat{I}_D-\vect{v}(u)\vect{v}(u)^\top)\vect{g}\|},\quad
\widetilde{\vect{z}}=\widehat{\vect{\gamma}}(u)+a\vect{r}(u),
\label{eq:sampling}
\end{equation}
where $a\sim\mathcal N(0,\sigma^2)$ independently. Code uses a $10^{-12}$ denominator floor for degenerate directions. Equation~\eqref{eq:sampling} therefore produces trajectory-centered samples with Gaussian normal displacement and uniform progression sampling. A nonuniform target progression density is not estimated from the $K$ shots.

\subsection{Observation-aware transfer to photon records}
\label{sec:obsaware}
The same coefficient-prior construction can be paired with an indirect observation model. In the photon experiment, let $\mathcal T_j$ denote the observed arrival times in a record of duration $T$ and set
\begin{equation}
\log D_j(t)=(1-t/T)w_{j0}+(t/T)w_{j1},\qquad
\vect{w}_j=(w_{j0},w_{j1})^\top\sim\mathcal N(\vect{\mu},\vect{\Lambda}).
\label{eq:photoncurve}
\end{equation}
This family uses observed clock time. Successful source fits under the event quasi-likelihood $\ell_{\rm ev}(\vect{w};\mathcal T_j)$ determine a mean and covariance, with an approximate inverse-Hessian correction and regularization (Appendix~\ref{app:photon}). This is not unbiased recovery of the latent source distribution.

For a target record, observation-aware HB-TPM (OA-HB-TPM) computes
\begin{equation}
\widehat{\vect{w}}_c=\argmin_{\vect{w}\in\mathcal B}
\left\{-\ell_{\rm ev}(\vect{w};\mathcal T_c)
+\tfrac12(\vect{w}-\widehat{\vect{\mu}})^\top
\widehat{\vect{\Lambda}}^{-1}(\vect{w}-\widehat{\vect{\mu}})\right\},
\label{eq:photonmap}
\end{equation}
where $\mathcal B=[\log 10,\log 220]^2$. A finite-state Markov-modulated Poisson surrogate and bounded optimization replace the conjugate vector update. Occupancy is estimated from counts and calibrated optics. Neither fitting stage receives true diffusion coefficients, positions, or molecular identities. We evaluate $D(t)$ and marginal endpoint intervals, not generated photon streams. Appendix~\ref{app:photon} specifies the surrogate and its approximations.

\subsection{Image-encoded periodic trajectories}
\label{sec:imageextension}
For cardiac images, an encoder maps each observed frame to a 16-dimensional mean and diagonal variance. A three-harmonic Fourier basis maps a Gaussian coefficient matrix to the full latent cycle. A convolutional decoder predicts left-ventricular mask probabilities. Target inference is heteroskedastic Gaussian conditioning on these encodings, with an image-derived circular phase offset. Unlike the vector plug-in fit, the encoder, decoder, phase head, and coefficient prior are trained jointly on source-patient masks. Thus ``observation-aware'' here denotes a learned latent observation model, not a likelihood for raw ultrasound pixels. Training combines segmentation, latent observation, coefficient-KL, cycle, and temporal losses. Appendix~\ref{app:ted} defines the objective, baseline variants, and numerical safeguards.

\section{Identifiability, shrinkage, and sampling stability}
\label{sec:theory}
Three statements connect the model to its recovery objective. The first identifies what finite observations leave unresolved, the second characterizes the conditional coefficient update, and the third bounds sampling error given an accurate curve. Their proofs use standard perturbation, Gaussian conditioning, and coupling arguments (Appendix~\ref{app:proofs}).

\begin{proposition}[Finite observations do not identify a smooth curve]
\label{prop:nonid}
Given any $C^2$ curve $\vect{\gamma}:[0,1]\to\R^D$, $D\geq1$, and finitely many coordinates $t_1,\ldots,t_K\in[0,1]$, there exists a $C^2$ curve with a different image that agrees with $\vect{\gamma}$ in a neighborhood of every $t_i$.
\end{proposition}
This unrestricted-function result shows why source-based structure is needed when a target provides only finitely many observations.

\begin{samepage}
\begin{proposition}[Conditional Gaussian shrinkage]
\label{prop:shrinkage}
Fix the coordinates, source hyperparameters, and one output dimension in the correctly specified model~\eqref{eq:hierarchy}, with $\vect{\Lambda}\succ0$ and $\sigma>0$. The posterior covariance and mean are Eq.~\eqref{eq:map}, and the mean is the unique conditional MAP. For a fixed probability measure $q$ on $[0,1]$ and $\mat{B}_q=\int\phiv(t)\phiv(t)^\top dq(t)$, its posterior integrated squared-error risk is
\begin{equation}
\E\left[\int\{\phiv(t)^\top(\widehat{\vect{w}}-\vect{w})\}^2dq(t)\,\middle|\,\vect{z}\right]
=\tr(\mat{B}_q\vect{\Sigma}).
\label{eq:bayesrisk}
\end{equation}
If $\mat{F}$ has full column rank, then $\vect{\Sigma}\prec\sigma^2(\mat{F}^\top\mat{F})^{-1}$. If it is rank deficient, the positive-definite prior still makes the update well posed.
\end{proposition}
\end{samepage}
The mechanism is especially transparent in prior-whitened coordinates. If $\mat{A}=\sigma^{-1}\mat{F}\vect{\Lambda}^{1/2}$ has singular values $s_b$, posterior variances along its right-singular directions are $(1+s_b^2)^{-1}$. Null directions retain unit prior variance and zero posterior mean displacement from the prior center. Thus when $K<p$, at least $p-K$ coefficient directions remain entirely prior-determined conditional on the indices.

\begin{proposition}[Sampling stability conditional on curve error]
\label{prop:tube}
Fix continuous curves $\widehat{\vect{\gamma}},\vect{\gamma}:[0,1]\to\R^D$ with $\sup_t\|\widehat{\vect{\gamma}}(t)-\vect{\gamma}(t)\|\leq\epsilon$. Couple $\widetilde{\vect{z}}=\widehat{\vect{\gamma}}(U)+\vect{\eta}$ to $\vect{z}^\star=\vect{\gamma}(U)$ using the same $U\in[0,1]$. If $\E\|\vect{\eta}\|\leq\tau$, then for every $L$-Lipschitz $f:\R^D\to\R$,
\begin{equation}
|\E f(\widetilde{\vect{z}})-\E f(\vect{z}^\star)|\leq L(\epsilon+\tau).
\label{eq:coupling}
\end{equation}
For every $\rho>\epsilon$,
$\Pr\{\dist(\widetilde{\vect{z}},\Gamma)>\rho\}\leq
\Pr\{\|\vect{\eta}\|>\rho-\epsilon\}$, where $\Gamma$ is the true curve image. If $\Gamma$ has reach $r_0>\rho$, points within this tube have a unique nearest point on $\Gamma$.
\end{proposition}
Here reach is the supremum of tube radii for which nearest-point projection is unique \citep{federer1959curvature}. For Eq.~\eqref{eq:sampling}, $\tau=\sigma\sqrt{2/\pi}$ away from degeneracies and the tail is bounded by $2\bar\Phi((\rho-\epsilon)/\sigma)$. The resulting stability bound is conditional on the curve approximation error $\epsilon$.

\section{Experiments}
\label{sec:experiments}
The experiments test recovery of missing variation, coefficient shrinkage, and adaptation under indirect observations. Tables~\ref{tab:main}--\ref{tab:ted} use task-specific comparators for vector distributions, diffusion curves, and cardiac masks. Their shared roles are target fitting, source transfer, and component ablation. Table~\ref{tab:datasets} summarizes datasets and information access. Earlier non-hierarchical image and spiral studies appear in Appendix~\ref{app:legacy}.

\begin{figure}[t]
\centering
\includegraphics[width=\linewidth]{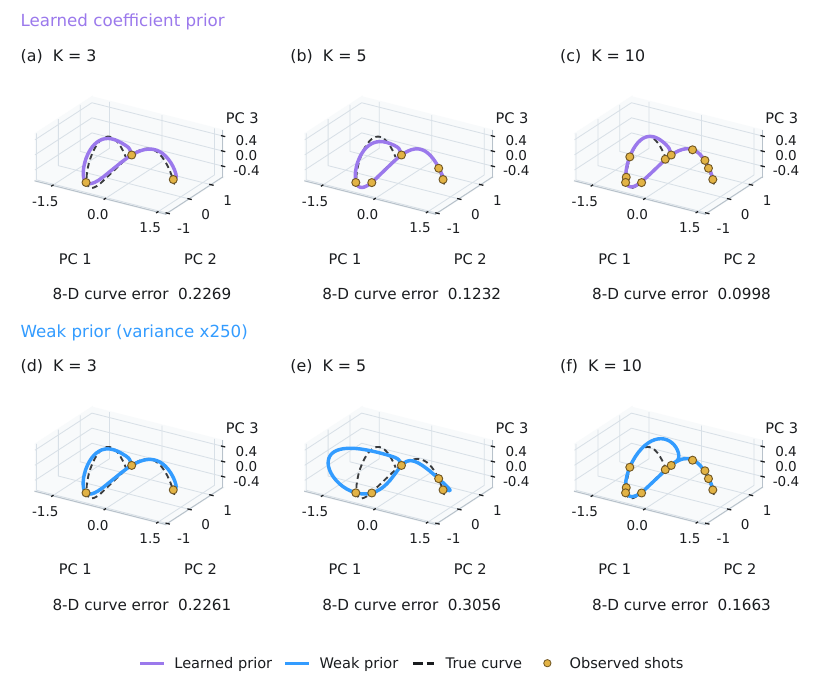}
\caption{Observations and prior strength in the first held-out target. Columns use nested $K=3,5,10$ shots (gold), with the learned prior above and its variance-inflated ($\times250$) ablation below. Dashed curves are truth. Panels share a source-only PCA map and limits. Annotations give symmetric curve error in eight dimensions. The learned prior improves recovery at $K=5,10$, while the weak prior has slightly lower error at $K=3$. Table~\ref{tab:main} reports aggregate distribution recovery.}
\label{fig:budgetprior3d}
\end{figure}

\subsection{Vector benchmark: protocol and information}
An eight-dimensional smooth trajectory family supplies 48 source tasks with 96 ordered observations each and 30 held-out targets with nested $K\in\{3,5,10\}$ unordered shots. Each target has 192 fresh oracle observations for evaluation. Target coordinates, simulator coefficients, and true curves are excluded from fitting. Truth-based projections and support metrics are evaluation-only (Appendix~\ref{app:implementation}).

Means and standard errors describe 30 targets conditional on one source cohort and base seed. The 90 task--budget comparisons are not independent tasks.

Baselines are Gaussian shrinkage, arbitrary-pair interpolation, PCA-ordered naive TPM, and target-trained and source-adapted VAEs. Both VAEs use source standardization. Naive TPM uses the source residual scale. Only HB-TPM variants use source progression coordinates. Thus comparisons combine source-information and modeling differences, rather than isolating architecture under identical information.

The primary metric is empirical sliced 2-Wasserstein distance (SWD) over 192 generated samples and 64 shared projection directions \citep{bonneel2015sliced}. Secondary metrics are energy distance, generated-to-curve and curve-to-generated distances, and symmetric curve error. All are lower-is-better. The curve distances are nearest-neighbor diagnostics, not probabilistic precision or credible-band coverage. Appendix~\ref{app:metrics} provides definitions and all-baseline visual comparisons.

\begin{table}[t]
\caption{\textbf{Synthetic 8-D trajectories: distribution recovery.} Sliced 2-Wasserstein distance (mean $\pm$ standard error over 30 targets, lower is better). Bold marks the lowest displayed means, not significance. Budgets are nested and uncertainty conditions on one source cohort. Only HB-TPM variants use source coordinates. Both VAEs use source standardization.}
\label{tab:main}
\centering
\small
\begin{tabular}{lccc}
\toprule
Method & $K=3$ & $K=5$ & $K=10$ \\
\midrule
Gaussian shrinkage & $0.2687 \pm 0.0137$ & $0.2089 \pm 0.0085$ & $0.1786 \pm 0.0056$ \\
Pairwise interpolation & $0.2837 \pm 0.0157$ & $0.2010 \pm 0.0117$ & $0.1601 \pm 0.0090$ \\
Naive TPM & $0.2910 \pm 0.0144$ & $0.1945 \pm 0.0121$ & $0.1539 \pm 0.0075$ \\
Target-trained VAE & $0.2762 \pm 0.0143$ & $0.2004 \pm 0.0100$ & $0.1783 \pm 0.0077$ \\
Source-adapted VAE & $0.1494 \pm 0.0037$ & $0.1340 \pm 0.0033$ & $0.1270 \pm 0.0033$ \\
\midrule
HB-TPM (normal) & $0.1180 \pm 0.0045$ & $0.0934 \pm 0.0047$ & $0.0739 \pm 0.0031$ \\
HB-TPM (weak prior) & $0.1426 \pm 0.0165$ & $0.1729 \pm 0.0306$ & $0.1273 \pm 0.0144$ \\
HB-TPM (isotropic) & $\mathbf{0.1175 \pm 0.0049}$ & $\mathbf{0.0926 \pm 0.0038}$ & $\mathbf{0.0636 \pm 0.0034}$ \\
\bottomrule
\end{tabular}
\end{table}

\subsection{Distribution recovery and mechanism checks}
The default HB-TPM sampler achieves mean SWD of $0.1180$, $0.0934$, and $0.0739$ at $K=3,5,10$ (Table~\ref{tab:main}), reductions of $21.0\%$, $30.2\%$, and $41.8\%$ relative to the source-adapted VAE. It has lower SWD on 28, 28, and 30 of the 30 paired targets, respectively. These comparisons support distribution recovery in this family, subject to the source-information difference described above.

Local fidelity and recovered extent separate interpolation from trajectory transfer. At $K=3$, pairwise interpolation yields lower generated-to-curve distance ($0.1411$ versus $0.2584$), while HB-TPM yields lower curve-to-generated distance ($0.2579$ versus $0.6517$) and SWD ($0.1180$ versus $0.2837$). Staying near a small observed segment can give good one-sided proximity while leaving much of the trajectory uncovered. Figure~\ref{fig:vectorextent} visualizes this distinction, complementing the full baseline comparison in Figure~\ref{fig:vectorcomparisons}. Joint assessment of fidelity and extent follows the broader rationale for quality-and-coverage evaluation \citep{kynkaanniemi2019improved}.

\paragraph{Prior strength and the gaps between observations.}
Multiplying coefficient variances by 250, with the source mean and other fitting choices unchanged, increases mean curve error from $0.2588/0.1717/0.0845$ to $0.3047/0.2957/0.1623$ at the three budgets. SWD also worsens. Figure~\ref{fig:budgetprior3d} shows what this ablation changes in the first target: the weak-prior curve at $K=5$ deviates between observed points, whereas the learned prior limits that deviation. At $K=3$ the weak prior is slightly better, so this example is not a uniform dominance claim. The aggregate evidence supports shrinkage within this family, not every prior strength or an unregularized estimator.

\paragraph{Normal versus isotropic perturbations.}
The isotropic ablation keeps the curve and matches expected squared noise radius using covariance $(\sigma^2/D)\mat{I}_D$. Its mean SWD is lower at every budget, most clearly at $K=10$ ($0.0636$ versus $0.0739$). Thus the stronger evidence is for curve estimation with shrinkage, not a benefit from normal-only perturbations. Separate sampling draws add Monte Carlo variation to this directional comparison.

\paragraph{An off-trajectory diagnostic.}
On full-dimensional Gaussian data, Gaussian shrinkage outperforms naive TPM in 82 of 90 comparisons. This warns against imposing a target-only curve on arbitrary distributions, but does not test HB-TPM under family shift: HB-TPM is absent from this diagnostic. The full protocol and results are in Appendix~\ref{app:control}.

\subsection{Photon benchmark: transfer through an observation model}
\label{sec:photonresults}
We simulate 40 source and 100 target records from one log-linear diffusion family. Thirty-nine successful source fits determine the prior. We compare three documented end-to-end estimators: target-only fitting uses no coefficient prior, the source mean uses no target photons, and OA-HB-TPM combines both. Target fits share events, bounds, calibrations, and quasi-likelihood, and each uses its specified initialization procedure. Table~\ref{tab:photon} retains all 100 targets, including nine target-only boundary fits. Bootstrap intervals condition on this source cohort.

\begin{table}[t]
\caption{\textbf{Synthetic photon arrivals: diffusion recovery.} 100 paired targets.
Mean relative curve RMSE is lower-is-better. Brackets are 95\% whole-record bootstrap intervals
conditional on one source cohort. All finite estimates, including nine
target-only boundary fits, are retained. Coverage averages 200 recorded
indicators for nominal 95\% marginal endpoint intervals, not whole curves.
The source mean uses no target photons. Success and coverage do not apply to
this point predictor. Bold marks the lowest mean error, most successful fits,
and coverage closest to 95\%, not statistical significance.}
\label{tab:photon}
\centering
\begin{tabular}{lrrrr}
\toprule
Method & Mean error & 95\% CI & Successful fits & Coverage \\
\midrule
Target-only event QMLE & 0.4378 & [0.3875, 0.4927] & 91/100 & 82\% \\
Source mean (post-hoc) & 0.2855 & [0.2568, 0.3163] & n/a & --- \\
OA-HB-TPM & \textbf{0.2180} & [0.1928, 0.2450] & \textbf{100/100} & \textbf{91\%} \\
\bottomrule
\end{tabular}
\end{table}

OA-HB-TPM reduces mean error by $50.2\%$ versus target-only fitting ($0.4378$ to $0.2180$) and by $23.6\%$ versus source-mean prediction ($0.2855$). The paired differences are $-0.2198$ (95\% interval $[-0.2632,-0.1800]$) and $-0.0675$ ($[-0.1067,-0.0292]$). It improves 88/100 and 60/100 targets, respectively. The reduction versus target-only fitting remains $47.8\%$ on the 91 jointly successful records. Adaptation therefore adds predictive value beyond the learned mean in this experiment. Calibration is a separate issue: nominal 95\% endpoint coverage is $91\%$ (versus $82\%$), and source endpoint correlation is estimated as $0.041$ rather than $0.650$. Appendix~\ref{app:photon} reports these diagnostics and the surrogate-model limitations.

\subsection{TED: sparse-frame cardiac reconstruction}
\label{sec:tedresults}
\paragraph{Protocol and comparisons.}
TED contains 98 annotated cardiac cycles \citep{painchaud2022temporal}. Five patient-disjoint test folds have separate validation and source-training patients. Sequences use 24 distinct $112\times112$ frames, with $K\in\{2,3,5,7,10\}$ observed images at uniform, random, or clustered times. Hidden images and target masks are excluded from inference. Table~\ref{tab:ted} averages six configurations over three training seeds and adds an OA setting with observation-noise scale 4.0. Replicates are averaged within patient--budget--pattern cells, followed by equal weighting of cells and seeds. Repeated seeds retain the same 98 patients.

\emph{Weak-prior adaptation} retains source-trained networks with a broad coefficient prior. \emph{Fixed source coefficients} retains image-derived phase but fixes coefficients. \emph{Latent-MSE training} replaces the observation loss while retaining Gaussian inference. Each system is trained separately.

\begin{table}[t]
\centering
\caption{\textbf{TED ultrasound: three-seed sparse-frame reconstruction.} 98 patients, five held-out folds. Replicates are averaged within patient--budget--pattern cells, then cells and training seeds receive equal weight. The last row uses observation-noise scale 4.0. Other rows retain the original protocol. Bold marks the best mean reconstruction score and coverage closest to nominal 95\%, not statistical significance. Coverage is pointwise soft-area coverage.}
\label{tab:ted}
\label{tab:ted_table3_final}
\vspace{6pt}
\small
\setlength{\tabcolsep}{4pt}
\begin{tabular}{lrrrr}
\toprule
Method & Dice $\uparrow$ & HD95 (px) $\downarrow$ & Area RMSE $\downarrow$ & 95\% coverage \\
\midrule
Weak-prior adaptation & 0.7454 & 10.0020 & 0.6926 & 43.6\% \\
Fixed source coefficients & 0.7472 & 9.8771 & 0.6622 & --- \\
HB-TPM (latent-MSE training) & \textbf{0.7519} & \textbf{9.4843} & \textbf{0.6357} & \textbf{63.6\%} \\
OA-HB-TPM (original) & 0.7461 & 9.9467 & 0.6728 & 39.3\% \\
OA-HB-TPM (isotropic prior) & 0.7461 & 9.9472 & 0.6727 & 39.1\% \\
OA-HB-TPM (no phase alignment) & 0.7462 & 9.9459 & 0.6726 & 39.3\% \\
OA-HB-TPM (noise scale 4.0) & 0.7461 & 9.9404 & 0.6716 & 42.7\% \\
\bottomrule
\end{tabular}
\end{table}

\paragraph{Reconstruction and calibration.}
HB-TPM with latent-MSE training achieves the best reported mean on all three reconstruction metrics: Dice $0.7519$, HD95 $9.4843$ pixels, and area RMSE $0.6357$. The seven rows compare complete configurations, so the result supports the latent-MSE HB-TPM configuration as a whole rather than isolating any single training or prior component.

Observation-noise scale 4.0 increases OA area coverage from $39.3\%$ to $42.7\%$ and lowers area RMSE from $0.6728$ to $0.6716$, with unchanged Dice at reported precision. Latent-MSE reaches $63.6\%$ coverage. All remain below nominal $95\%$. Noise scaling therefore yields a modest coverage improvement, not calibrated uncertainty.

\section{Discussion and conclusion}
\label{sec:limits}
HB-TPM makes missing-structure recovery explicit: ordered sources learn a trajectory family, sparse targets adapt it, and a sampler distributes mass around the recovered curve. Vector ablations support coefficient shrinkage as the central mechanism, and photon records demonstrate useful observation-aware adaptation. On TED, latent-MSE training is the strongest tested mean reconstruction configuration. The modest effect of noise scaling separates reconstruction quality from uncertainty calibration.

\paragraph{Scope and next comparisons.}
The model assumes comparable source progression. Each simulation uses one source cohort, and TED uses one dataset with three training seeds. GP/PACE/neural-process comparisons, frozen-encoder tests, and external cohorts would extend the evidence (Appendices~\ref{app:literature} and \ref{app:tedinterpretation}). Patient-level rerun records and scale-selection metadata are needed to quantify uncertainty around the updated cardiac means.

Learning a distribution of plausible trajectories supplies structure for sparse-target recovery. Explicit coefficient inference makes the contributions of source information, target observations, and the sampling rule separately testable.

\label{sec:mainend}

\FloatBarrier
\subsection*{AI use statement}
Generative AI tools (ChatGPT/Codex) assisted with conceptual development, experiment design, implementation, mathematical formulations and proof writing, result interpretation, literature search, manuscript editing, and figure preparation. Synthetic data were generated by explicit numerical simulators. Vector and photon summaries were checked against experiment-level outputs, and the vector benchmark was reproduced. The three-seed cardiac table uses the reported aggregate results, with its verification scope specified in Appendix~\ref{app:ted}. The authors are responsible for the correctness and originality of all content, including AI-assisted text, code, proofs, and figures.
\subsection*{Ethics statement}
The vector, photon, and spiral experiments are simulated. The cardiac study uses existing, publicly available, de-identified TED ultrasound sequences and annotations \citep{painchaud2022temporal}. No new participants were recruited. The supplementary package supplies derived evaluation records and code, not patient image arrays. Archived digit/face illustrations in Appendix~\ref{app:legacyimages} are not used to infer personal attributes or support clinical claims. A misspecified source prior or image representation can bias predictions, and the observed interval undercoverage precludes reliance on the reported uncertainty for clinical decisions. The study does not evaluate clinical deployment or downstream task performance.
\subsection*{Reproducibility statement}
Algorithm~\ref{alg:hb} specifies vector inference. Sections~\ref{sec:obsaware} and \ref{sec:imageextension} describe the observation-aware variants. Appendices~\ref{app:proofs}--\ref{app:ted} give proofs, settings, metrics, and diagnostics, and Appendix~\ref{app:reproduce} summarizes the reproduction protocol. Vector and photon means and paired contrasts were recomputed from experiment-level results. Table~\ref{tab:ted} reports updated three-seed cardiac aggregate means, with the aggregation and evaluation protocol specified in Appendix~\ref{app:ted}. Appendix~\ref{app:ted} distinguishes these means from the earlier single-seed diagnostics and specifies the records needed for paired uncertainty and calibration checks. Appendix~\ref{app:legacy} separates the reproduced spiral pilot from archival image panels that cannot be independently regenerated.
\bibliography{references_iclr2027}
\bibliographystyle{iclr2027_conference}
\clearpage
\appendix
\section*{Appendix}
\section{Proofs and scope of the theoretical statements}
\label{app:proofs}
\subsection{Non-identifiability without a restricted family}
\begin{proof}[Proof of Proposition~\ref{prop:nonid}]
Because $\{t_1,\ldots,t_K\}$ is finite, there is a nonempty open interval $I$ with closure inside $(0,1)$ that contains none of these coordinates. Choose a nonzero $C^\infty$ bump function $b$ whose support is contained in $I$, a nonzero vector $\vect{v}\in\R^D$, and a scalar $c\ne0$. Define
\[
\widetilde{\vect{\gamma}}(t)=\vect{\gamma}(t)+c b(t)\vect{v}.
\]
This curve is $C^2$ and equals $\vect{\gamma}$ outside the compact support of $b$, including a relative neighborhood in $[0,1]$ of every observed coordinate. To ensure a different curve image, not merely a different parameterization, take $t_0$ with $b(t_0)\ne0$ and set $R=\max_{t\in[0,1]}\|\vect{\gamma}(t)\|<\infty$. Choose $c$ so that $|c b(t_0)|\|\vect{v}\|>2R$. The reverse triangle inequality then gives
\[
\|\widetilde{\vect{\gamma}}(t_0)\|
\geq |c b(t_0)|\|\vect{v}\|-\|\vect{\gamma}(t_0)\|>R.
\]
Thus $\widetilde{\vect{\gamma}}(t_0)$ is outside the original curve image. The observations identify neither the parameterized curve nor its image in this unrestricted class. The construction applies even when the coordinates are supplied, and therefore also when they are not supplied.
\end{proof}
No injectivity, curvature bound, or finite-basis assumption is imposed in this proposition. The claim must not be strengthened into an impossibility result for every restricted trajectory class, nor into a quantitative minimax rate without further assumptions.

\subsection{Conditional Gaussian inference}
\begin{proof}[Proof of Proposition~\ref{prop:shrinkage}]
Suppress the output index. The conditional negative log posterior, up to a constant independent of $\vect{w}$, is
\[
\frac{1}{2\sigma^2}\|\vect{z}-\mat{F}\vect{w}\|^2
+\frac12(\vect{w}-\vect{w}_0)^\top\vect{\Lambda}^{-1}(\vect{w}-\vect{w}_0).
\]
Its Hessian is $\mat{P}=\sigma^{-2}\mat{F}^\top\mat{F}+\vect{\Lambda}^{-1}\succ0$, including when $\mat{F}$ is rank deficient. With $\vect{b}=\sigma^{-2}\mat{F}^\top\vect{z}+\vect{\Lambda}^{-1}\vect{w}_0$, completing the square gives
\[
\tfrac12\vect{w}^{\top}\mat{P}\vect{w}-\vect{w}^{\top}\vect{b}
=\tfrac12(\vect{w}-\mat{P}^{-1}\vect{b})^{\top}\mat{P}(\vect{w}-\mat{P}^{-1}\vect{b})
-\tfrac12\vect{b}^{\top}\mat{P}^{-1}\vect{b}.
\]
The posterior is therefore a proper Gaussian with covariance $\vect{\Sigma}=\mat{P}^{-1}$ and mean $\widehat{\vect{w}}=\mat{P}^{-1}\vect{b}$. Strict convexity establishes uniqueness of this conditional MAP. All conditioning here holds the coordinates and hyperparameters fixed. It is not a posterior integrating their uncertainty.

Under this posterior, $\vect{w}-\widehat{\vect{w}}$ has mean zero and covariance $\vect{\Sigma}$. For each $t$,
\[
\E\bigl[\{\phiv(t)^\top(\widehat{\vect{w}}-\vect{w})\}^2\mid\vect{z}\bigr]
=\phiv(t)^\top\vect{\Sigma}\phiv(t).
\]
Integration and the trace identity yield Eq.~\eqref{eq:bayesrisk}. Interchange of expectation and integration follows from nonnegativity. Finiteness follows because the smooth basis is bounded on $[0,1]$ and the covariance is finite.

If $\mat{F}$ has full column rank, set $\mat{H}=\sigma^{-2}\mat{F}^\top\mat{F}\succ0$. Then
\[
\vect{\Sigma}=\mat{H}^{-1/2}
\bigl(\mat{I}+\mat{H}^{-1/2}\vect{\Lambda}^{-1}\mat{H}^{-1/2}\bigr)^{-1}\mat{H}^{-1/2}
\prec\mat{H}^{-1}.
\]
The inner added matrix is positive definite, so every eigenvalue of $(\mat{I}+\mat{H}^{-1/2}\vect{\Lambda}^{-1}\mat{H}^{-1/2})^{-1}$ is strictly below one. Congruence by $\mat{H}^{-1/2}$ preserves the strict positive-definite ordering and proves the covariance comparison.
\end{proof}

The comparison concerns posterior uncertainty, not fixed-parameter estimation error. For example, with scalar prior $w\sim\mathcal N(0,1)$ and observation $z\mid w\sim\mathcal N(w,1)$, the posterior mean is $z/2$. At fixed truth $w=5$, its frequentist squared-error risk is $1/4+25/4=6.5$, whereas the risk of $z$ is 1. A strong but incorrect prior can therefore increase target estimation error despite reducing conditional posterior covariance.

\paragraph{Which directions are learned from the target?}
Use the symmetric positive-definite square root of $\vect{\Lambda}$. Let $\vect{h}=\vect{\Lambda}^{-1/2}(\vect{w}-\vect{w}_0)$,
$\vect{y}=\sigma^{-1}(\vect{z}-\mat{F}\vect{w}_0)$, and
$\mat{A}=\sigma^{-1}\mat{F}\vect{\Lambda}^{1/2}$. Then
$\vect{h}\sim\mathcal N(\vect{0},\mat{I}_p)$ and
$\vect{y}\mid\vect{h}\sim\mathcal N(\mat{A}\vect{h},\mat{I}_K)$.
Completing the square once more gives
\begin{equation}
\widehat{\vect{h}}=(\mat{I}_p+\mat{A}^\top\mat{A})^{-1}\mat{A}^\top\vect{y},\qquad
\operatorname{Cov}(\vect{h}\mid\vect{y})=(\mat{I}_p+\mat{A}^\top\mat{A})^{-1}.
\label{eq:whitened}
\end{equation}
Take an orthonormal eigenbasis $\{\vect{v}_b\}_{b=1}^p$ of $\mat{A}^\top\mat{A}$, with eigenvalues $s_b^2\geq0$ (including zeros). Equation~\eqref{eq:whitened} implies
\[
\operatorname{Var}(\vect{v}_b^\top\vect{h}\mid\vect{y})=(1+s_b^2)^{-1},\qquad
\vect{v}_b^\top\widehat{\vect{h}}=
\frac{\vect{v}_b^\top\mat{A}^\top\vect{y}}{1+s_b^2}.
\]
If $s_b=0$, then $\mat{A}\vect{v}_b=\vect{0}$, so the mean displacement is zero and the variance remains one. Since $\operatorname{rank}(\mat{A})\leq K$, there are at least $p-K$ such directions when $K<p$. This establishes the interpretation used in the main text. The directions depend on the fixed design and prior. They are not generally individual Bernstein coefficients, and may change when the projection indices are updated.

For completeness, at a fixed coefficient truth $\vect{w}^\star$ with
$\vect{z}=\mat{F}\vect{w}^\star+\vect{e}$ and
$\vect{e}\sim\mathcal N(\vect{0},\sigma^2\mat{I}_K)$, let
$\mat{H}=\sigma^{-2}\mat{F}^\top\mat{F}$ and
$\vect{b}=\vect{\Sigma}\vect{\Lambda}^{-1}(\vect{w}_0-\vect{w}^\star)$.
The frequentist bias and covariance of the posterior mean are respectively
$\vect{b}$ and $\vect{\Sigma}\mat{H}\vect{\Sigma}$. Consequently,
\begin{equation}
\E_{\vect{e}}\!\left[\int\{\phiv(t)^\top(\widehat{\vect{w}}-\vect{w}^\star)\}^2dq(t)\right]
=\vect{b}^\top\mat{B}_q\vect{b}
+\tr(\mat{B}_q\vect{\Sigma}\mat{H}\vect{\Sigma}).
\label{eq:fixedrisk}
\end{equation}
To verify this identity, substitute the observation equation into Eq.~\eqref{eq:map} and use
$\vect{\Sigma}(\vect{\Lambda}^{-1}+\mat{H})=\mat{I}_p$.
The error becomes $\vect{b}+\sigma^{-2}\vect{\Sigma}\mat{F}^\top\vect{e}$.
Its cross term has zero expectation. This explicitly separates source-prior bias from observation variance. It is not a bound for the complete latent-index estimator.

\paragraph{A decomposition separating index and prior errors.}
Let $\vect{t}^\star$ denote the true target coordinates, let $\mat{W}^\star$ be any fixed coefficient approximation to the true curve, and let $\widehat{\mat{W}}(\vect{t};\theta)$ denote Eq.~\eqref{eq:map} for hyperparameters $\theta$. Write $\widehat\theta$ for source estimates and $\theta_0$ for a fixed reference. Define
\begin{align*}
\mat{E}_{\rm ind}&=\widehat{\mat{W}}(\widehat{\vect{t}};\widehat\theta)-\widehat{\mat{W}}(\vect{t}^\star;\widehat\theta),\\
\mat{E}_{\rm src}&=\widehat{\mat{W}}(\vect{t}^\star;\widehat\theta)-\widehat{\mat{W}}(\vect{t}^\star;\theta_0),\\
\mat{E}_{\rm coef}&=\widehat{\mat{W}}(\vect{t}^\star;\theta_0)-\mat{W}^\star,\qquad
\epsilon_{\rm fam}=\sup_t\|\mat{W}^{\star\top}\phiv(t)-\vect{\gamma}(t)\|.
\end{align*}
The coefficient error telescopes to $\mat{E}_{\rm ind}+\mat{E}_{\rm src}+\mat{E}_{\rm coef}$. With $C_\phi=\sup_t\|\phiv(t)\|$ and $\|\mat{E}^{\top}\phiv(t)\|\leq\|\mat{E}\|_F\|\phiv(t)\|$, the triangle inequality gives
\begin{equation}
\sup_t\|\widehat{\vect{\gamma}}(t)-\vect{\gamma}(t)\|
\leq C_\phi\bigl(\|\mat{E}_{\rm ind}\|_F+\|\mat{E}_{\rm src}\|_F+\|\mat{E}_{\rm coef}\|_F\bigr)+\epsilon_{\rm fam}.
\label{eq:decomposition}
\end{equation}
Proposition~\ref{prop:shrinkage} describes the conditional coefficient component under its model. It does not bound the other components. Equation~\eqref{eq:decomposition} is a deterministic decomposition, not a statistical convergence rate. For the Bernstein basis used here, nonnegative basis entries sum to one, so $C_\phi\leq1$.

\subsection{Sampling stability}
\begin{proof}[Proof of Proposition~\ref{prop:tube}]
The continuous curves are bounded on $[0,1]$. Together with $\E\|\vect{\eta}\|<\infty$, this ensures finite expectations for Lipschitz $f$. The curves are held fixed (or conditioned on the fitting data). No independence of $U$ and $\vect{\eta}$ is required. For the specified coupling, pointwise,
\[
\|\widetilde{\vect{z}}-\vect{z}^\star\|
\leq\|\widehat{\vect{\gamma}}(U)-\vect{\gamma}(U)\|+\|\vect{\eta}\|
\leq\epsilon+\|\vect{\eta}\|.
\]
Lipschitz continuity gives
$|f(\widetilde{\vect{z}})-f(\vect{z}^\star)|\leq L(\epsilon+\|\vect{\eta}\|)$.
Taking expectations proves Eq.~\eqref{eq:coupling}. Also $\vect{\gamma}(U)\in\Gamma$, so
$\dist(\widetilde{\vect{z}},\Gamma)\leq\epsilon+\|\vect{\eta}\|$.
The asserted tail bound follows by event inclusion. The curve image $\Gamma$ is compact, so a nearest point exists. Reach and $\rho<r_0$ ensure its uniqueness for $\dist(\vect{x},\Gamma)\leq\rho$. This is uniqueness of a spatial point, not of its parameter value: distinct coordinates can map to the same point. No normal-direction assumption is needed for these inequalities.
\end{proof}
When $\vect{\eta}=a\vect{r}(U)$ with $\|\vect{r}(U)\|=1$, its norm is $|a|$. Thus $\E\|\vect{\eta}\|=\sigma\sqrt{2/\pi}$ and
$\Pr(\|\vect{\eta}\|>x)=2\bar\Phi(x/\sigma)$ for $x\geq0$.
The denominator floor can only reduce the norm of a projected direction, so these remain upper bounds in degenerate cases. A Gaussian-radius implementation has no deterministic tube-support guarantee. Truncating its radius would define a different sampler and is not used in the reported benchmark.

The coupling uses the same progression law for fitted and true curves. Hausdorff proximity of curve images alone is insufficient: the uniform distribution on $[0,1]$ and the distribution with density $2t$ have the same support but means $1/2$ and $2/3$. To compare with a noisy oracle rather than a noiseless curve, one must also account for the oracle's observation noise. For example, adding an oracle perturbation $\vect{\eta}^\star$ to the coupled reference yields the general bound $L(\epsilon+\E\|\vect{\eta}\|+\E\|\vect{\eta}^\star\|)$ by the triangle inequality. Sharper bounds require a specified noise coupling.

\paragraph{Connection to distributional error.}
If $\E\|\vect{\eta}\|^2\leq\tau_2^2$, the same coupling and the $L^2$ triangle inequality give
\[
W_2\bigl(\mathcal L(\widetilde{\vect{z}}),\mathcal L(\vect{z}^\star)\bigr)
\leq\{\E\|\widetilde{\vect{z}}-\vect{z}^\star\|^2\}^{1/2}
\leq\epsilon+\tau_2.
\]
Here $\mathcal L$ denotes distribution and $W_2$ the 2-Wasserstein distance. Bounded curves and the noise assumption ensure finite second moments. Projection onto any unit direction is 1-Lipschitz. Averaging squared projected distances against a probability measure on the unit sphere and taking a square root therefore gives the same bound for population sliced 2-Wasserstein distance. For the normal-radius sampler, $\tau_2\leq\sigma$, including denominator-floor degeneracies. A noisy oracle adds its own second-moment radius to this general bound. This result concerns population distributions with the same progression law. It neither bounds finite-sample SWD automatically nor supplies a data-derived value of $\epsilon$.

\subsection{Alternating minimization on the finite grid}
With fixed positive effective variances, each assignment step minimizes the squared-error term over grid positions independently for the $K$ observations. The prior term is unchanged. Each coefficient step minimizes the strictly convex conditional objective. Thus, in exact arithmetic,
\[
\mathcal J(\mat{W}^{r+1},\vect{t}^{r+1})
\leq\mathcal J(\mat{W}^{r},\vect{t}^{r+1})
\leq\mathcal J(\mat{W}^{r},\vect{t}^{r}).
\]
If consecutive assignments coincide, the next coefficient fit is the same unique minimizer. With deterministic tie-breaking, this pair is a blockwise fixed point, so stopping is justified. The cap of 20 iterations can stop earlier. Monotonicity establishes neither a global optimum nor the correct latent coordinates. Per-target iteration histories were not retained, so convergence before the cap has not been established for every target.

\subsection{Fixed-prior limits at a singular design}
\label{app:flatlimit}
For the reduction in Appendix~\ref{app:imagespecialization}, fix $\mat{F}$, $\sigma>0$, and prior $\mathcal N(\vect{w}_0,\alpha^{-1}\mat{I}_p)$ with $\alpha>0$. Let $r=\operatorname{rank}(\mat{F})$ and write a singular-value decomposition
$\mat{F}=\sum_{b=1}^{r}s_b\vect{u}_b\vect{v}_b^{\top}$ with $s_b>0$. Extend the right singular vectors to an orthonormal basis of $\R^p$. In this basis, the posterior formulas give
\begin{align*}
\widehat{\vect{w}}_\alpha
&=\vect{w}_0+\sum_{b=1}^{r}\frac{s_b}{s_b^2+\sigma^2\alpha}\,
\vect{v}_b\vect{u}_b^{\top}(\vect{z}-\mat{F}\vect{w}_0),\\
\vect{\Sigma}_\alpha
&=\sum_{b=1}^{r}\frac{\sigma^2}{s_b^2+\sigma^2\alpha}\vect{v}_b\vect{v}_b^{\top}
+\frac1\alpha\sum_{b=r+1}^{p}\vect{v}_b\vect{v}_b^{\top}.
\end{align*}
The first identity follows by writing the update as $\vect{w}_0+(\mat{F}^{\top}\mat{F}+\sigma^2\alpha\mat{I}_p)^{-1}\mat{F}^{\top}(\vect{z}-\mat{F}\vect{w}_0)$. Taking $\alpha\downarrow0$ gives
$\mat{F}^{\dagger}\vect{z}+(\mat{I}_p-\mat{F}^{\dagger}\mat{F})\vect{w}_0$.
For $\vect{w}_0=\vect{0}$ this is the minimum-norm least-squares solution, even if $r<p$. However, every null-direction variance is $1/\alpha$. At zero precision the likelihood is constant along these directions, so its integral over $\R^p$ is infinite when $r<p$: the flat-prior posterior is improper despite the convergent point estimate. Empty sums cover $r=0$ or $r=p$.

\paragraph{Numerical checks.}
The supplement includes seeded tests of the Gaussian posterior identities, prior-whitened null directions, fixed-truth risk decomposition, Bernstein basis and derivative, and non-increasing alternating objective. Edge-case checks cover rank-zero designs, prior-retaining flat limits, the deterministic curve-error decomposition, dependent sampling noise, and degenerate tangents. These are implementation checks, not replacements for the proofs or additional empirical benchmark results.

\section{Experimental settings and vector implementation}
\label{app:implementation}
\begin{table}[htbp]
\caption{\textbf{Datasets and experimental roles.} The first three rows evaluate hierarchical transfer. The remaining rows retain diagnostics and earlier non-hierarchical studies. Different observations require different prediction targets and comparators. Image-study summaries are not pooled with the current benchmarks.}
\label{tab:datasets}
\centering
\small
\setlength{\tabcolsep}{4pt}
\renewcommand{\arraystretch}{1.16}
\begin{tabular}{@{}>{\raggedright\arraybackslash}p{0.16\linewidth}>{\raggedright\arraybackslash}p{0.27\linewidth}>{\raggedright\arraybackslash}p{0.25\linewidth}>{\raggedright\arraybackslash}p{0.26\linewidth}@{}}
\toprule
Dataset / setting & Size and split & Estimator-visible input & Prediction and comparison \\
\midrule
Synthetic 8-D trajectories & 48 sources $\times$ 96 ordered points. 30 held-out targets, $K=3,5,10$. & Source coordinates and vectors for HB-TPM. Target vectors have no coordinates. & Sample distribution and curve recovery. Gaussian, pairwise, naive TPM, VAEs, and HB ablations. \\
Synthetic photon arrivals & 40 source and 100 target records. Prior fitted from 39 successful sources. & Event times, duration, and calibrated observation parameters. Clock time is observed. & Diffusion trajectory. Target-only event QMLE, source mean, and OA-HB-TPM. \\
TED ultrasound \citep{painchaud2022temporal} & 98 patients, five patient-disjoint test folds, three training seeds. 24 frames per cycle. $K=2,3,5,7,10$. & Sparse target images and times. Source training uses masks. Hidden images and target masks are excluded. & Hidden-frame masks and area curves. Six original configurations plus OA observation-noise scale 4.0. \\
\midrule
Synthetic Gaussian control & 30 repeats per budget, $K=3,5,10$. No HB-TPM run. & Sparse target vectors. Naive TPM also uses an oracle noise scale. & Distribution recovery by Gaussian shrinkage versus naive TPM. Off-trajectory diagnostic only. \\
Synthetic spiral pilot & Three classes. $K=3,5$ per class. 2,100 test points, one seed per cell. & Target-only labeled shots. No source trajectory prior. & Classification by seven recorded prototype methods. Reproduced, but not compute-matched. \\
MNIST / face image studies \citep{lecun1998gradient,liu2015faceattributes} & Earlier figure and summary records. Exact splits, identities, and checkpoints are incomplete. & Encoded static images under the earlier target-only workflow. & Qualitative geometry and interpolation, reported smoothness and classification. Not tests of hierarchical transfer. \\
\bottomrule
\end{tabular}
\end{table}

\paragraph{Why baseline sets differ.}
A vector sampler, an event-likelihood estimator, and an image-to-mask model solve different statistical problems. A VAE fitted to vectors is not directly a diffusion estimator, and event QMLE is not a mask decoder. We therefore compare each method within its observation model. The vector study compares support estimators and generative models, the photon study separates source-only prediction from target-only and prior-regularized estimation, and TED tests prior and observation-model components within a shared architectural family. These roles explain Tables~\ref{tab:main}--\ref{tab:ted}, but do not make the designs information-matched: vector baselines differ in their use of source coordinates, and TED variants are trained separately. The experiments support within-task conclusions, not a universal cross-domain ranking. A shared future baseline family would pair source-trained GP, PACE, or neural-process priors with each task's observation model.

\FloatBarrier
\subsection{Synthetic trajectory family and information separation}
Let $s=2t-1$ and define the simulator-only basis
\[
\vect{\psi}(t)=(1,s,\sin(\pi s),\cos(\pi s),\sin(2\pi s),\cos(2\pi s))^\top.
\]
A fixed $8\times8$ orthogonal matrix $\mat{Q}$ is obtained by QR decomposition of a standard-Gaussian matrix with seed 20260822. Before multiplying by $\mat{Q}$, the $6\times8$ base coefficient matrix has only five nonzero entries (using one-based indices): $(2,1)=1.20$, $(3,2)=0.38$, $(4,2)=0.72$, $(5,3)=0.16$, and $(6,3)=0.32$. Each task adds independent Gaussian coefficient perturbations with row standard deviations $(0.10,0.07,0.09,0.09,0.055,0.055)$ to the rotated base matrix. The task curve is the resulting coefficient matrix transposed times $\vect{\psi}(t)$.

Coordinates are uniform on $[0,1]$ and sorted for source tasks only. Simulator observations perturb the curve normally using a unit projected Gaussian direction and an independent radius $\mathcal N(0,0.035^2)$. Source fitting accesses only observed coordinates and vectors. Target fitting accesses only its observed vectors and the learned prior. Simulator coefficients, target coordinates, and the oracle basis are not passed to the estimator. The true curve is evaluated on 401 grid points for support metrics. Figure~\ref{fig:recovery} uses a PCA projection derived from this true curve solely as an evaluation visualization.

\subsection{HB-TPM hyperparameters and computational cost}
The fitted basis is $\phi_b(t)=\binom{9}{b}t^b(1-t)^{9-b}$, $b=0,\ldots,9$. Source ridge penalty is $\lambda_s=10^{-6}$, and diagonal prior variances use the sample variance across 48 fitted tasks with denominator 47 and floor $2.5\times10^{-4}$. The source residual scale is
\[
\widehat\sigma=\operatorname{clip}\left(\frac{\operatorname{median}_{j,i}\|\vect{z}_{ji}-\widehat{\mat{W}}_j^\top\phiv(t_{ji})\|}{0.6745},0.015,0.08\right).
\]
The fitted value is $0.03525945$. Its use as both a radial sampling scale and a working isotropic likelihood scale is a modeling approximation. Coefficient variances are additionally floored at $10^{-7}$ after multiplication by the ablation factor, and working noise variance is floored at $10^{-6}$. Target projection uses $G=401$ equally spaced positions and at most 20 iterations. Sampling draws uniform coordinates continuously, unlike the discrete inference grid.

Ignoring precomputable basis evaluations, one target iteration requires $O(GpD+KGD+Kp^2+D p^3)$ arithmetic for curve evaluation, assignment, a shared Gram matrix, and dense coordinatewise solves. The implementation redundantly forms the Gram matrix in the coordinate loop, adding a factor $D$ to that term. The expression above describes the natural cached implementation. No GPU is required for this small NumPy benchmark. This is an operation count, not a measured runtime comparison.

\subsection{Baseline details}
\paragraph{Gaussian shrinkage.}
Let $\bar{\vect{z}}$ be the target mean, $\mat{S}$ its sample covariance with divisor $K-1$, and $\bar v=\max\{\tr(\mat{S})/D,10^{-5}\}$. For $K>1$, use $\alpha=\min\{0.55,D/(D+K)\}$ and draw from
\[
\mathcal N\bigl(\bar{\vect{z}},(1-\alpha)\mat{S}+\alpha\bar v\mat{I}_D+10^{-6}\mat{I}_D\bigr).
\]
This is not the Gaussian maximum-likelihood covariance estimator, and it does not estimate a diagonal covariance even in the diagonal-Gaussian control. For $K=1$, the covariance before the $10^{-6}$ jitter is $10^{-3}\mat{I}_D$. This case is not used in the experiment.

\paragraph{Pairwise interpolation and naive TPM.}
The former samples two target indices independently with replacement and a uniform mixing weight. It therefore includes self-pairs and does not restrict interpolation to nearest neighbors. Naive TPM projects centered targets onto the first right singular vector, sorts them by that projection, forms normalized cumulative chord-length knots, and linearly interpolates each ambient coordinate onto 401 points. Its generated samples select grid points uniformly and add normal-direction perturbations with the source residual scale. No spline optimization, source mean curve, or target posterior is used.

\paragraph{VAEs.}
Both VAEs use a tanh encoder with 28 hidden units, separate two-dimensional Gaussian mean and log-variance outputs, and a tanh decoder with 28 hidden units and linear eight-dimensional output. The log variance is clipped to $[-6,3]$. Training uses a half-squared reconstruction loss plus $\beta=0.035$ times latent KL divergence, with reparameterized Gaussian noise. Adam uses moments $(0.9,0.999)$, numerical offset $10^{-8}$, and elementwise gradient clipping at $[-5,5]$. Generated outputs are decoder means for standard-normal latent draws, without additional decoder observation noise.

Both baselines standardize using pooled-source means and coordinate standard deviations floored at $0.08$. Source pretraining takes 2400 updates with batch size 128 and learning rate $0.003$. Target-trained VAE parameters start from scratch and use 700 updates at $0.004$. Source adaptation starts from the pretrained parameters, uses 180 updates at $0.0015$, and adds an anchor gradient $0.025(\theta-\theta_{\rm source})$. Target minibatches contain $K$ draws with replacement: although the requested batch size is $\max(K,8)$, the implementation caps it at the number of target observations. Thus target-trained parameters do not imply source-free preprocessing.

\paragraph{Ablations.}
The weak-prior variant multiplies all source coefficient variances by 250. It keeps the source mean, working noise, grid, iteration cap, and normal sampler. The isotropic variant reuses the default fitted coefficients and replaces normal perturbations by $\mathcal N(\vect{0},(\sigma^2/D)\mat{I}_D)$. Both perturbation rules have expected squared radius $\sigma^2$ away from numerical degeneracy. Their generated samples are independent draws from successive states of the same random-number generator. Common random numbers were not used for this ablation.

\FloatBarrier
\subsection{Source-to-target algorithm and companion views}
\begin{figure}[htbp]
\centering
\includegraphics[width=\linewidth]{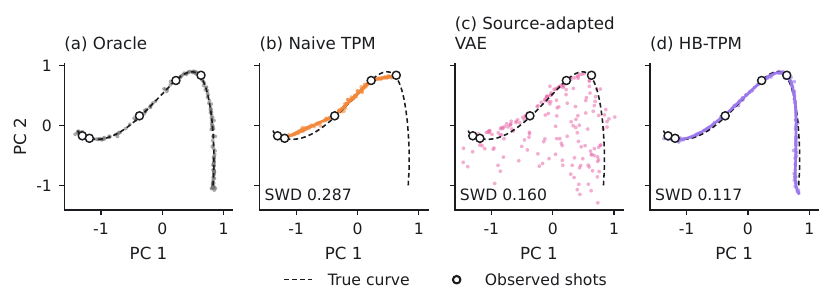}
\caption{Distribution recovery for the first held-out five-shot target. Panels share an evaluation-only PCA projection fitted to the true eight-dimensional curve. Dashed lines show that curve and black-rimmed circles mark the same observations. Other points are oracle or generated samples. HB-TPM recovers unobserved regions in this example. SWD annotations use the full eight-dimensional space. Table~\ref{tab:main} gives the aggregate comparison across 30 targets.}
\label{fig:recovery}
\end{figure}

Algorithm~\ref{alg:hb} states the implemented alternating updates. Figure~\ref{fig:framework} separates source fitting, target adaptation, and generation. Unlike the data-driven views in Figures~\ref{fig:sourcefamily3d} and \ref{fig:budgetprior3d}, this workflow is schematic. Figure~\ref{fig:recovery} retains the four-method sample comparison for the same held-out target.

\begin{algorithm}[htbp]
\caption{Empirical-Bayes HB-TPM}
\label{alg:hb}
\begin{algorithmic}[1]
\REQUIRE Ordered source pairs, target matrix $\mat{Z}$, basis $\phiv$, grid $\mathcal G$, sample count $M$.
\STATE Fit source coefficient matrices by Eq.~\eqref{eq:source}. Estimate $\mat{W}_0$, diagonal $\vect{\Lambda}_d$, and residual scale $\sigma$.
\STATE Initialize $\mat{W}\leftarrow\mat{W}_0$ and previous indices to unset.
\FOR{$r=1,\ldots,20$}
\STATE Assign each $\widehat t_i$ by Eq.~\eqref{eq:projection}, breaking ties by first grid index.
\STATE Update every $\vect{w}_d$ by Eq.~\eqref{eq:map}.
\IF{all assigned indices equal those of the preceding iteration}
\STATE Stop.
\ENDIF
\STATE Store current indices.
\ENDFOR
\STATE Generate $M$ samples using Eq.~\eqref{eq:sampling} with the fitted $\mat{W}$.
\ENSURE Fitted curve, grid assignments, and synthetic target vectors.
\end{algorithmic}
\end{algorithm}

\FloatBarrier

\FloatBarrier
\subsection{Projection and example-selection protocol}
\label{app:3dviews}
Figures~\ref{fig:sourcefamily3d} and \ref{fig:budgetprior3d} visualize the synthetic vector benchmark, not the image embeddings in Appendix~\ref{app:legacyrepresentation}. Both use the first held-out target, also shown in Figure~\ref{fig:recovery}. No search over targets or selection by method ranking is performed. Source observations and the first target are reconstructed from the benchmark seed and simulator. The five-shot observations, true curve, fitted HB-TPM curve, and all six prior-ablation curve errors agree with the saved record.

A single PCA map is fitted to all $48\times96=4608$ observed source vectors. Its first three components retain $89.1\%$ of source variance. Target observations and simulator truth are excluded from this fit. PCA is used only for display: all estimation and errors remain eight-dimensional. Axis limits and viewing angles are shared within each figure. Figure~\ref{fig:sourcefamily3d}(b) displays all 48 source ridge fits, while its target panel reuses the 192 saved HB-TPM samples. The spread of source curves represents between-task variation, not a target credible region.

Figure~\ref{fig:budgetprior3d} uses the benchmark's nested shot sets and weak-prior multiplier of 250. It retains the same prior mean, working noise, projection grid, and iteration cap. The learned-prior errors at $K=3,5,10$ are $0.2269$, $0.1232$, and $0.0998$, versus $0.2261$, $0.3056$, and $0.1663$ with the weak prior. Thus the first target illustrates improved recovery at the two larger budgets but not uniform superiority. The quantitative conclusions use all 30 targets, rather than this example. Projected proximity can conceal differences in omitted coordinates.

Figure~\ref{fig:recovery} uses a different, evaluation-only two-dimensional PCA map fitted to the true target curve. It compares sample distributions rather than coefficient-prior strength. The map is shared across its four panels and never supplied to an estimator. Its SWD annotations are computed in the full eight-dimensional space.

\FloatBarrier
\section{Metrics and complete numerical summaries}
\label{app:metrics}
\suppressfloats[t]
\subsection{All-baseline visual comparisons}
\begin{figure}[htbp]
\centering
\includegraphics[width=\linewidth]{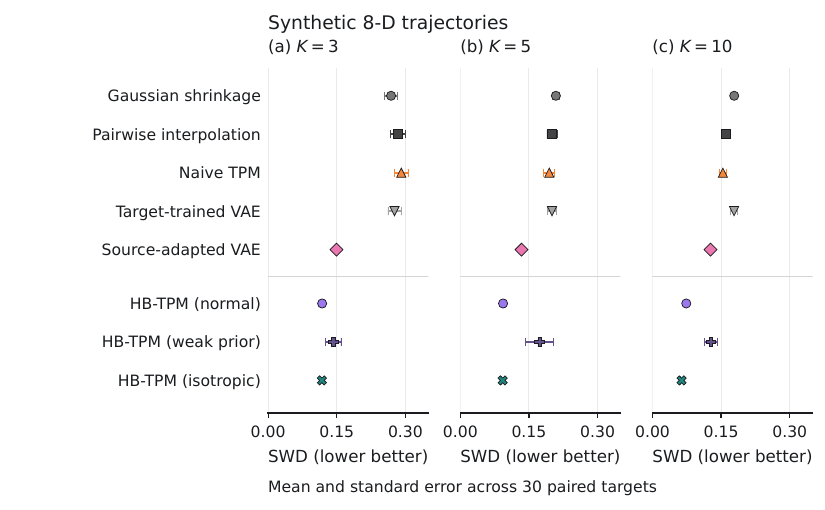}
\caption{\textbf{Synthetic 8-D trajectories: all baselines and ablations.} Points show mean SWD, with one-standard-error whiskers over 30 targets at each budget. Lower is better. All three panels use the same axis limits and method order. Budgets are nested, so these are not 90 independent targets. Estimates condition on one source cohort. The line separates comparison methods from HB-TPM variants, not information-matched groups. Only the latter receive source progression coordinates. Isotropic sampling has the lowest mean at all three budgets.}
\label{fig:vectorcomparisons}
\end{figure}

\begin{figure}[htbp]
\centering
\includegraphics[width=\linewidth]{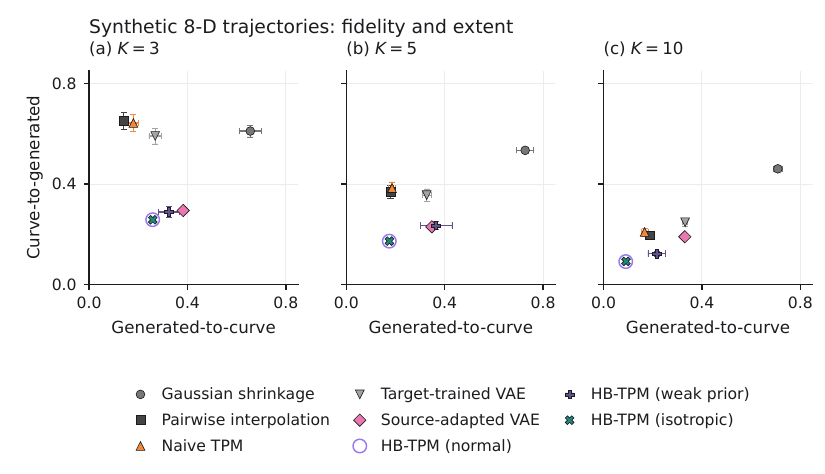}
\caption{\textbf{Synthetic 8-D trajectories: local fidelity versus recovered extent.} Each point is a method's mean generated-to-curve distance (horizontal) and curve-to-generated distance (vertical), with one-standard-error whiskers across 30 targets. Lower is better in both directions. At $K=3$, interpolation stays close to the curve but covers less of it than the HB-TPM variants. The larger open purple circle makes normal HB-TPM visible where it nearly coincides with the isotropic marker, without shifting either point. These nearest-neighbor diagnostics are not probabilistic precision, recall, or uncertainty coverage.}
\label{fig:vectorextent}
\end{figure}

Figure~\ref{fig:vectorcomparisons} shows the complete SWD comparison without combining targets across budgets. Figure~\ref{fig:vectorextent} separates proximity to the curve from coverage of its extent. Both redraw the same records as the numerical tables, not additional experiments.
\FloatBarrier
\subsection{Definitions and uncertainty}
For equally sized generated and oracle samples $\{\vect{x}_m\}_{m=1}^M$ and $\{\vect{y}_m\}_{m=1}^M$, and unit directions $\vect{\theta}_\ell$, the implemented SWD is
\[
\widehat{\operatorname{SW}}_2=\left[\frac1{LM}\sum_{\ell=1}^L\sum_{m=1}^M
\bigl(\langle\vect{\theta}_\ell,\vect{x}\rangle_{(m)}-
\langle\vect{\theta}_\ell,\vect{y}\rangle_{(m)}\bigr)^2\right]^{1/2},
\quad L=64,\ M=192.
\]
Here $(m)$ indexes sorted projected samples. The reported energy statistic is the biased empirical V-statistic
\[
\widehat{\mathcal E}=\frac{2}{M^2}\sum_{m,n}\|\vect{x}_m-\vect{y}_n\|
-\frac1{M^2}\sum_{m,n}\|\vect{x}_m-\vect{x}_n\|
-\frac1{M^2}\sum_{m,n}\|\vect{y}_m-\vect{y}_n\|.
\]
Let $\mathcal C$ be the 401-point true-curve grid and $\widehat{\mathcal C}$ an estimated-curve grid. Define
\begin{align*}
d_{\rm gen\to curve}&=\frac1M\sum_m\min_{\vect{c}\in\mathcal C}\|\vect{x}_m-\vect{c}\|,\\
d_{\rm curve\to gen}&=\frac1{|\mathcal C|}\sum_{\vect{c}\in\mathcal C}\min_m\|\vect{c}-\vect{x}_m\|,\\
d_{\rm sym}(\widehat{\mathcal C},\mathcal C)&=\frac12\left[
\frac1{|\widehat{\mathcal C}|}\sum_{\widehat{\vect{c}}\in\widehat{\mathcal C}}\min_{\vect{c}\in\mathcal C}\|\widehat{\vect{c}}-\vect{c}\|
+\frac1{|\mathcal C|}\sum_{\vect{c}\in\mathcal C}\min_{\widehat{\vect{c}}\in\widehat{\mathcal C}}\|\vect{c}-\widehat{\vect{c}}\|\right].
\end{align*}
The first two quantities are distances, not probabilities or coverage fractions. A small value of one does not imply a small value of the other. These discrete averages are neither Hausdorff distance nor the uniform parameter-aligned curve error in Proposition~\ref{prop:tube}.

Each table cell is the arithmetic mean over 30 targets, with sample standard deviation divided by $\sqrt{30}$ as its standard error. No confidence interval across independently trained source families is available. Each target's budgets are nested, so cross-budget summaries require clustering by target if used inferentially. Tables~\ref{tab:secondary}--\ref{tab:curve} retain all methods rather than selecting favorable metric--method pairs.

\begin{table}[htbp]
\caption{\textbf{Synthetic 8-D trajectories: energy statistic.} Mean $\pm$ standard error across 30 targets, lower is better. Bold marks the lowest displayed means (rounding ties included), not significance.}
\label{tab:secondary}
\centering
\small
\begin{tabular}{lccc}
\toprule
Method & $K=3$ & $K=5$ & $K=10$ \\
\midrule
Gaussian shrinkage & $0.3235 \pm 0.0380$ & $0.1938 \pm 0.0184$ & $0.1351 \pm 0.0090$ \\
Pairwise interpolation & $0.4095 \pm 0.0446$ & $0.2226 \pm 0.0239$ & $0.1353 \pm 0.0143$ \\
Naive TPM & $0.4429 \pm 0.0430$ & $0.2113 \pm 0.0229$ & $0.1194 \pm 0.0100$ \\
Target-trained VAE & $0.4003 \pm 0.0407$ & $0.2256 \pm 0.0209$ & $0.1709 \pm 0.0133$ \\
Source-adapted VAE & $0.1273 \pm 0.0062$ & $0.1026 \pm 0.0044$ & $0.0885 \pm 0.0037$ \\
HB-TPM (normal) & $0.0877 \pm 0.0074$ & $0.0519 \pm 0.0045$ & $0.0265 \pm 0.0022$ \\
HB-TPM (weak prior) & $0.1426 \pm 0.0352$ & $0.1634 \pm 0.0547$ & $0.0682 \pm 0.0143$ \\
HB-TPM (isotropic) & $\mathbf{0.0856 \pm 0.0074}$ & $\mathbf{0.0515 \pm 0.0042}$ & $\mathbf{0.0224 \pm 0.0023}$ \\
\bottomrule
\end{tabular}
\end{table}

\begin{table}[htbp]
\caption{\textbf{Synthetic 8-D trajectories: generated-to-curve distance.} Mean $\pm$ standard error across 30 targets, lower is better. Bold marks the lowest displayed means (rounding ties included), not significance. This one-sided distance measures proximity, not support coverage. Read together with Table~\ref{tab:coverage} and SWD.}
\label{tab:precision}
\centering
\small
\begin{tabular}{lccc}
\toprule
Method & $K=3$ & $K=5$ & $K=10$ \\
\midrule
Gaussian shrinkage & $0.6553 \pm 0.0456$ & $0.7271 \pm 0.0347$ & $0.7086 \pm 0.0172$ \\
Pairwise interpolation & $\mathbf{0.1411 \pm 0.0135}$ & $0.1800 \pm 0.0108$ & $0.1881 \pm 0.0065$ \\
Naive TPM & $0.1799 \pm 0.0193$ & $0.1859 \pm 0.0145$ & $0.1671 \pm 0.0133$ \\
Target-trained VAE & $0.2687 \pm 0.0252$ & $0.3269 \pm 0.0191$ & $0.3317 \pm 0.0103$ \\
Source-adapted VAE & $0.3824 \pm 0.0109$ & $0.3476 \pm 0.0068$ & $0.3304 \pm 0.0058$ \\
HB-TPM (normal) & $0.2584 \pm 0.0104$ & $\mathbf{0.1747 \pm 0.0085}$ & $\mathbf{0.0904 \pm 0.0056}$ \\
HB-TPM (weak prior) & $0.3252 \pm 0.0432$ & $0.3646 \pm 0.0644$ & $0.2164 \pm 0.0359$ \\
HB-TPM (isotropic) & $0.2587 \pm 0.0110$ & $0.1752 \pm 0.0088$ & $0.0912 \pm 0.0056$ \\
\bottomrule
\end{tabular}
\end{table}

\begin{table}[htbp]
\caption{\textbf{Synthetic 8-D trajectories: curve-to-generated distance.} Mean $\pm$ standard error across 30 targets, lower is better. Bold marks the lowest displayed means (rounding ties included), not significance.}
\label{tab:coverage}
\centering
\small
\begin{tabular}{lccc}
\toprule
Method & $K=3$ & $K=5$ & $K=10$ \\
\midrule
Gaussian shrinkage & $0.6111 \pm 0.0232$ & $0.5342 \pm 0.0124$ & $0.4605 \pm 0.0078$ \\
Pairwise interpolation & $0.6517 \pm 0.0345$ & $0.3698 \pm 0.0261$ & $0.1953 \pm 0.0147$ \\
Naive TPM & $0.6432 \pm 0.0341$ & $0.3834 \pm 0.0249$ & $0.2088 \pm 0.0156$ \\
Target-trained VAE & $0.5914 \pm 0.0318$ & $0.3557 \pm 0.0246$ & $0.2478 \pm 0.0171$ \\
Source-adapted VAE & $0.2944 \pm 0.0111$ & $0.2295 \pm 0.0079$ & $0.1899 \pm 0.0064$ \\
HB-TPM (normal) & $0.2579 \pm 0.0119$ & $\mathbf{0.1723 \pm 0.0093}$ & $\mathbf{0.0909 \pm 0.0065}$ \\
HB-TPM (weak prior) & $0.2887 \pm 0.0218$ & $0.2351 \pm 0.0166$ & $0.1230 \pm 0.0100$ \\
HB-TPM (isotropic) & $\mathbf{0.2575 \pm 0.0121}$ & $0.1725 \pm 0.0094$ & $0.0916 \pm 0.0064$ \\
\bottomrule
\end{tabular}
\end{table}

\begin{table}[htbp]
\caption{\textbf{Synthetic 8-D trajectories: symmetric curve error.} Mean $\pm$ standard error across 30 targets, lower is better. Bold marks the lowest displayed means (rounding ties included), not significance. A dash denotes a method without an explicit curve output.}
\label{tab:curve}
\centering
\small
\begin{tabular}{lccc}
\toprule
Method & $K=3$ & $K=5$ & $K=10$ \\
\midrule
Gaussian shrinkage & --- & --- & --- \\
Pairwise interpolation & --- & --- & --- \\
Naive TPM & $0.4170 \pm 0.0167$ & $0.2838 \pm 0.0114$ & $0.1825 \pm 0.0101$ \\
Target-trained VAE & --- & --- & --- \\
Source-adapted VAE & --- & --- & --- \\
HB-TPM (normal) & $\mathbf{0.2588 \pm 0.0115}$ & $\mathbf{0.1717 \pm 0.0090}$ & $\mathbf{0.0845 \pm 0.0064}$ \\
HB-TPM (weak prior) & $0.3047 \pm 0.0300$ & $0.2957 \pm 0.0355$ & $0.1623 \pm 0.0222$ \\
HB-TPM (isotropic) & $\mathbf{0.2588 \pm 0.0115}$ & $\mathbf{0.1717 \pm 0.0090}$ & $\mathbf{0.0845 \pm 0.0064}$ \\
\bottomrule
\end{tabular}
\end{table}

\FloatBarrier

\subsection{Full-dimensional Gaussian diagnostic}
\label{app:control}
For each of 30 control repeats, draw a mean $\vect{\mu}\sim\mathcal N(\vect{0},0.3^2\mat{I}_8)$. Observations are independent Gaussian vectors with coordinate standard deviations linearly spaced from $0.02$ to $0.08$. At each $K$, draw fresh target observations and an independent oracle evaluation sample. Unlike the main benchmark, shot sets are not nested across budgets.

Only the covariance-shrinkage Gaussian and naive TPM are evaluated. Naive TPM's perturbation radius uses the true mean coordinate standard deviation ($0.05$), an oracle-provided scale rather than a fitted source scale. The SWD calls also consume successive states of one metric generator, so the two methods use different projection directions. Table~\ref{tab:control} reports this limited comparison.

\begin{table}[htbp]
\caption{\textbf{Synthetic 8-D Gaussian data: off-trajectory diagnostic.} SWD is lower-is-better. Bold marks the lower mean within each row. HB-TPM was not evaluated. Methods use different SWD projection draws. Naive TPM uses an oracle noise scale.}
\label{tab:control}
\centering
\small
\begin{tabular}{cccc}
\toprule
$K$ & Gaussian shrinkage & Naive TPM & Gaussian wins \\
\midrule
3 & $\mathbf{0.0347 \pm 0.0017}$ & $0.0406 \pm 0.0012$ & 26/30 \\
5 & $\mathbf{0.0290 \pm 0.0012}$ & $0.0341 \pm 0.0011$ & 29/30 \\
10 & $\mathbf{0.0213 \pm 0.0008}$ & $0.0254 \pm 0.0009$ & 27/30 \\
\bottomrule
\end{tabular}
\end{table}

The Gaussian estimator has lower SWD in $26/30$, $29/30$, and $27/30$ comparisons at $K=3,5,10$. This comparison shows why a target-only curve need not suit full-dimensional data. Because HB-TPM was not evaluated, it does not measure hierarchical-transfer robustness or family-shift detection. Such a study requires direct HB-TPM evaluation, common metric projections, matched observations, and nuisance estimates obtained without oracle access.

\FloatBarrier
\clearpage
\section{Illustration of sampling geometry}
\label{app:geometry}
\begin{figure}[htbp]
\centering
\includegraphics[width=\linewidth]{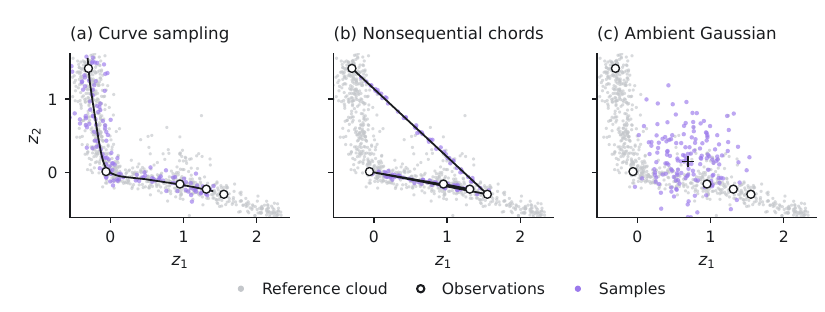}
\caption{Three sampling geometries in a synthetic two-dimensional example. Gray points form a constructed reference cloud. Outlined circles denote five fixed observations and purple points are synthetic samples. Left: perturbing a prescribed smooth curve illustrates trajectory-aligned sampling, not curve estimation from the five points. Middle: a fixed nonsequential ordering connects the observations by chords away from much of the cloud. Right: an isotropic Gaussian centered at their mean (cross) allocates samples away from the curved concentration of points. This is a geometric illustration, not empirical MNIST evidence or a quantitative comparison of fitted methods.}
\label{fig:syntheticgeometry}
\end{figure}

Figure~\ref{fig:syntheticgeometry} explains the motivation for modeling a curved target support before drawing synthetic samples. It is a constructed example, separate from all three experiments. The gray cloud is generated around a prescribed planar curve with additional synthetic scatter. The five outlined observations and the chord ordering are prespecified for this illustration. No MNIST images, learned embedding, source-prior estimation, or target curve fitting are used. The left panel presupposes a curve and therefore cannot establish that sparse observations identify it.

The reference cloud illustrates high-density geometry rather than compact probabilistic support: its Gaussian perturbations are unbounded. The drawing provides neither credible regions nor error bounds. Figure~\ref{fig:recovery} and Tables~\ref{tab:main}--\ref{tab:ted} provide the actual benchmark evidence.
\FloatBarrier
\clearpage
\section{Observation-aware photon experiment}
\label{app:photon}
This experiment evaluates transfer of a two-coefficient diffusion trajectory through an indirect observation model. Acquisition times are observed. The target is $D(t)$, not a molecular path or generated photon stream.

\subsection{Data generation and information separation}
For each independent record $j$, draw
\[
\vect{w}_j\sim\mathcal N\left(
\begin{bmatrix}\log45\\\log95\end{bmatrix},
\begin{bmatrix}0.18^2&0.65(0.18)(0.22)\\0.65(0.18)(0.22)&0.22^2\end{bmatrix}
\right).
\]
The diffusion curve is $D_j(t)=\exp\{(1-t/T)w_{j0}+(t/T)w_{j1}\}$.
Diffusion is expressed in $\mu{\rm m}^2/{\rm s}$ before taking logarithms, and $T=0.08\,{\rm s}$. Source and target coefficients are independent draws from this same distribution. The fixed mean endpoints $45$ and $95$ are exponentiated log means, not arithmetic means of the log-normal diffusion coefficients. Forty sources and 100 targets are generated under base seed 2026083001.

Each record contains 12 independently diffusing molecules, initially uniform in a periodic box of size $1.5\times1.5\times4.0\,\mu{\rm m}^3$. The hierarchy is
\begin{align}
d\vect{X}_{ij}(t)&=\sqrt{2D_j(t)}\,d\vect{B}_{ij}(t),\nonumber\\
h(x,y,z)&=\exp\{-2(x^2+y^2)/w_{xy}^2-2z^2/w_z^2\},\nonumber\\
\lambda_j(t)&=b+A_j\sum_{i=1}^{12}h\{\vect{X}_{ij}(t)\},
\label{eq:photonphysics}
\end{align}
with $w_{xy}=0.25\,\mu{\rm m}$, $w_z=1.25\,\mu{\rm m}$, and $b=1000\,{\rm s}^{-1}$. Molecular brightness $A_j$ is $60000\,{\rm s}^{-1}$ for sources and $25000\,{\rm s}^{-1}$ for targets. These are different-brightness records, not paired thinnings of the same molecular paths. No blinking, bleaching, or detector artifacts are added.

The simulator uses a $10^{-5}\,{\rm s}$ grid. Gaussian Brownian increments have coordinate variance $2\int_{t_k}^{t_{k+1}}D_j(s)\,ds$ and are wrapped periodically. The photon rate is evaluated at the left endpoint and held constant within each step. Counts are Poisson with mean $\lambda_j(t_k)\Delta t$. Conditional on each count, event times are uniform in that step and then sorted. This is exact photon generation conditional on the stepwise-constant rate, not exact observation of a continuous Brownian path. Step-halving convergence has not been evaluated.

Estimator-visible data consist of arrival times, record duration, and calibrated PSF widths, brightness, and background. The true diffusion curve, latent positions, molecular identities, and conditional intensity are not arguments to the fitting API. The supplied runner fits sources and targets before joining truth for scoring. Runtime provenance limitations are documented in Appendix~\ref{app:photonprovenance}.

\subsection{Finite-state event quasi-likelihood}
The fitted observation model differs from the Brownian simulator. Let $M(t)\in\{0,\ldots,12\}$ be a surrogate occupancy process, let $\kappa=w_z/w_{xy}=5$, and define
\[
q_{\rm eff}=A_j/2^{3/2},\qquad
\widehat n=\operatorname{clip}\left\{\frac{N_j/T-b}{q_{\rm eff}},0.02,2.0\right\},
\qquad
\mu(t)=\frac{4D_j(t)}{w_{xy}^2}\left(1+\frac{1}{2\kappa^2}\right).
\]
If the un-clipped occupancy estimate is nonpositive, the fit fails. The generator has upward rate $\widehat n\mu(t)$ for states below 12, downward rate $m\mu(t)$ from state $m>0$, and diagonal entries equal to minus the outgoing rate sum. At the upper boundary no upward transition is permitted. The photon-rate matrix is
\[
\mat{R}=\diag\{b+q_{\rm eff}m:m=0,\ldots,12\}.
\]
The initial distribution is the normalized Poisson distribution with mean $\widehat n$, truncated to these states. The inferred occupancy is a surrogate effective occupancy, not the simulated count of 12 molecules in the box.

Divide the record into 16 equal intervals and evaluate $D_j$ at their midpoints. In interval $r$, the generator $\mat{Q}_r(\vect{w})$ is frozen. A row filter $\vect{\alpha}$ propagates over a no-event gap $\Delta$ as
\[
\vect{\alpha}\leftarrow\vect{\alpha}
\exp\{(\mat{Q}_r(\vect{w})-\mat{R})\Delta\}.
\]
At an event, it is multiplied by $\mat{R}$. The filter is normalized at events and interval boundaries, with log normalizers accumulated into $\ell_{\rm ev}$. The final no-event interval contributes its survival probability. The filter is carried across refinement boundaries without resetting to stationarity.

These operations evaluate the likelihood of the declared midpoint-frozen, truncated surrogate, up to floating-point computation. They approximate a smoothly time-varying surrogate and provide a quasi-likelihood for the Brownian--Gaussian-PSF observations. No claim of an exact marginalized Brownian likelihood is made. Refinement and state-truncation sensitivity have not been established.

\subsection{Empirical prior, target fitting, and uncertainty}
Source fits maximize $\ell_{\rm ev}$ over $\mathcal B=[\log10,\log220]^2$. For each successful source fit, $\widehat{\vect{w}}_j$ is its coefficient estimate and $\mat{C}_j$ its regularized inverse-Hessian covariance. With $J_s=39$ successful fits, define
\begin{align}
\widehat{\vect{\mu}}&=J_s^{-1}\sum_{j=1}^{J_s}\widehat{\vect{w}}_j,\nonumber\\
\mat{S}&=(J_s-1)^{-1}\sum_{j=1}^{J_s}
(\widehat{\vect{w}}_j-\widehat{\vect{\mu}})
(\widehat{\vect{w}}_j-\widehat{\vect{\mu}})^\top
-J_s^{-1}\sum_{j=1}^{J_s}\mat{C}_j,\nonumber\\
\mat{S}_+&=\mat{U}\diag\{\max(\lambda_1,0.0025),\max(\lambda_2,0.0025)\}\mat{U}^\top,\nonumber\\
\widehat{\vect{\Lambda}}&=0.8\mat{S}_++0.2\diag(\diag(\mat{S}_+)),
\qquad \mat{S}=\mat{U}\diag(\lambda_1,\lambda_2)\mat{U}^\top.
\label{eq:photonprior}
\end{align}
Unlike the vector implementation, this is a full $2\times2$ covariance. Subtracting $\mat{C}_j$ is an approximate measurement-error correction. Source fit bias, covariance error, and selection of successful sources remain uncorrected. This is not a jointly optimized marginal-likelihood hierarchy.

Both target methods use bounded L-BFGS-B, at most 80 iterations, \texttt{ftol}$=10^{-9}$, \texttt{gtol}$=10^{-5}$, and at most 30 line-search steps. Target-only fitting initializes at the midpoint of the log bounds. OA-HB-TPM initializes at the learned prior mean clipped to the bounds. Thus the comparison differs in both regularization and initialization, not solely in a penalty evaluated from an identical start.

Central differences with step $0.002$ estimate the objective Hessian. Its eigenvalues are floored at $10^{-4}$ before inversion. With covariance $\widehat{\mat{C}}$, the nominal marginal endpoint intervals are
\[
\left[\exp\{\widehat w_d-1.96\sqrt{\widehat C_{dd}}\},
\exp\{\widehat w_d+1.96\sqrt{\widehat C_{dd}}\}\right],\quad d=0,1.
\]
These Laplace-style intervals omit occupancy and source-hyperparameter uncertainty, do not apply a quasi-likelihood sandwich correction, and are not simultaneous trajectory bands. Proposition~\ref{prop:shrinkage}'s exact conjugate calculation does not apply to this surrogate model.

A successful fit requires optimizer success, finite estimates and objective, neither log endpoint within $10^{-4}$ of a search bound, and no occupancy clipping. Information-eigenvalue regularization is recorded as a diagnostic but does not by itself set the success flag to false. The nine unsuccessful target-only fits all reach a diffusion search boundary and retain finite estimates.

\subsection{Metrics, paired analysis, and diagnostic results}
For target $j$, the relative curve error is
\[
e_j=\left[\frac1{201}\sum_{\ell=0}^{200}
\left\{\frac{\widehat D_j(\ell T/200)-D_j(\ell T/200)}
{D_j(\ell T/200)}\right\}^2\right]^{1/2}.
\]
The main summary is $100^{-1}\sum_j e_j$, not a pooled root-mean-square across records. Log endpoint RMSE is the square root of the mean of both endpoint squared log errors over records. Endpoint coverage averages the two marginal coverage indicators per record. Time points and photons are not counted as independent replicates.

We use 20000 percentile bootstrap resamples of paired target records, seed 20260917. Each resample retains the same records across methods, including each record's two coverage indicators. The resulting uncertainty is conditional on the single source cohort and does not include variability from relearning the prior. The relative reduction is the ratio-of-means statistic $1-\bar e_{\rm HB}/\bar e_{\rm ref}$, not the average of per-record percentage improvements.

The target-only and OA-HB-TPM log endpoint RMSEs are $0.6953$ and $0.3091$, respectively. Median relative curve errors are $0.4157$ and $0.1850$. The $50.2\%$ reduction in mean curve error has a paired 95\% interval of $[44.9\%,55.2\%]$. On the 91 jointly successful records, mean errors are $0.4303$ and $0.2245$, with paired difference interval $[-0.2514,-0.1631]$. These conditional-on-success summaries supplement, rather than replace, the all-record comparison.

The post-hoc source-mean predictor uses Eq.~\eqref{eq:photoncurve} with $\vect{w}=\widehat{\vect{\mu}}$ for every target. Target truth is used only for scoring. Its mean error is $0.2855$, versus $0.2180$ after target adaptation, with reduction interval $[11.1\%,34.6\%]$. No target observations or target-driven choice of this predictor's parameters is involved. It is a post-hoc diagnostic, not a separately fitted target model.

\begin{figure}[t]
\centering
\includegraphics[width=\linewidth]{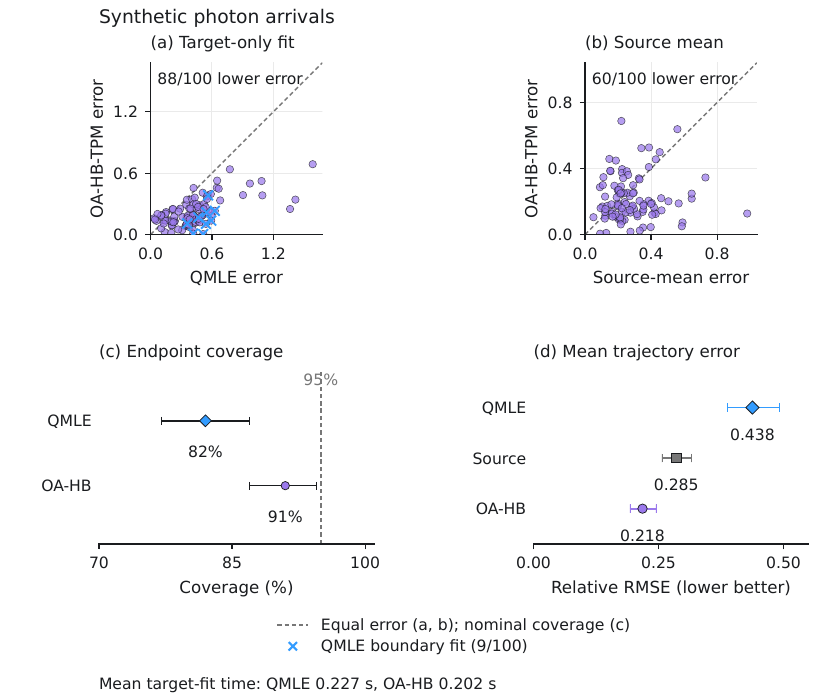}
\caption{\textbf{Synthetic photon arrivals: paired error, coverage, and aggregate performance.} (a--b) All 100 targets are shown, with relative trajectory RMSE on both axes (lower is better). Points below equality favor OA-HB-TPM. Blue crosses identify the nine target-only boundary fits, all retained. (c) Empirical coverage of nominal 95\% marginal endpoint intervals. (d) Mean trajectory error, including the source-mean predictor. Whiskers in (c--d) are 95\% intervals from 20000 whole-record bootstrap resamples, conditional on the learned prior. Coverage uses recorded endpoint-inclusion indicators, not reconstructed bounds or simultaneous curve coverage. Reported mean target-fit times exclude source-prior estimation and are descriptive measurements, not a controlled speed comparison.}
\label{fig:photondiagnostics}
\end{figure}

Source and target photon counts span 685--1554 and 352--653, respectively. Mean counts are $1098.9$ and $492.7$. Mean target-fitting times are $0.227$ seconds for event QMLE and $0.202$ seconds for OA-HB-TPM, excluding source learning, simulation, and overall overhead. These descriptive timings do not establish a hardware-independent speed advantage. Estimation uses SciPy on CPU.

Of the 200 nominal endpoint intervals, $164$ cover truth for QMLE and $182$ for OA-HB-TPM. For OA-HB-TPM, 82 records have both endpoints covered and 18 have one covered. Neither $182/200$ nor $82/100$ is simultaneous coverage of $D(t)$ for all $t$. Only coverage summaries were retained, not interval endpoints or per-target covariance matrices, preventing independent reclassification from interval bounds.

\begin{table}[htbp]
\caption{\textbf{Synthetic photon arrivals: source-prior diagnostics.} Exponentiated log means are not arithmetic diffusion means. Learned quantities derive from 39 successful source fits. Columns compare parameters with estimates, not competing methods, so no best score is designated.}
\label{tab:photonprior}
\centering
\begin{tabular}{lrr}
\toprule
Quantity & Generating family & Learned prior\\
\midrule
$\exp(\mu_0)$ ($\mu{\rm m}^2/{\rm s}$) & 45.00 & 63.50\\
$\exp(\mu_1)$ ($\mu{\rm m}^2/{\rm s}$) & 95.00 & 88.67\\
Log-start variance & 0.0324 & 0.1532\\
Log-end variance & 0.0484 & 0.0992\\
Endpoint correlation & 0.650 & 0.041\\
\bottomrule
\end{tabular}
\end{table}
The raw corrected covariance eigenvalues are $0.0984$ and $0.1540$, both above the $0.0025$ floor. Thus covariance clipping is not the explanation for this run's prior distortion. Mean target log contrast is $0.7609$ in truth and $0.4847$ after OA-HB-TPM fitting, indicating attenuation of the diffusion change. These discrepancies could reflect the surrogate likelihood, occupancy plug-in, finite information, or numerical approximation. The present comparisons cannot distinguish their contributions.

\subsection{Reproducibility and remaining validation}
\label{app:photonprovenance}
Record-level checks confirm the numbers of source and target observations, pairing, sorted in-range arrival times, and photon counts. Trajectory errors can be recomputed from fitted coefficients and separate evaluation truth. These checks establish data integrity and arithmetic consistency, not an independent rerun of the photon experiment.

The available outputs support coefficient-error rescoring and coverage-indicator reaggregation. They do not include version-linked execution provenance, full latent simulator diagnostics, interval endpoints, or per-record posterior covariances. Coverage therefore cannot be independently reclassified from its bounds. The experiment establishes a within-family prediction comparison, with numerical convergence, interval calibration, and shift robustness requiring further evaluation.

Next tests should retain complete uncertainty and execution metadata, compare two-stage empirical Bayes and covariance ablations on the same events, verify Brownian-step and likelihood-refinement stability, repeat source cohorts, and test matched photon budgets and explicit family shifts. Hyperparameters or calibration corrections chosen after inspecting these target truths require new held-out evaluation. The separate cardiac experiment is described in Appendix~\ref{app:ted}. It does not validate this photon observation model.

\FloatBarrier
\clearpage
\clearpage
\section{TED cardiac reconstruction: protocol and diagnostic results}
\label{app:ted}
This study tests whether source-trained trajectory structure supports reconstruction of left-ventricular cavity masks from sparse raw ultrasound frames. TED provides 98 fully annotated apical four-chamber sequences drawn from CAMUS \citep{painchaud2022temporal}. We use all 98 sequences, with one sequence identifier per patient. Table~\ref{tab:ted} reports the updated three-training-seed means. Figure~\ref{fig:teddiagnostics} retains the original single-seed budget diagnostics and is labeled separately. The prediction targets are hidden-frame masks and full-cycle area trajectories.

\subsection{Preprocessing and patient separation}
Each sequence is sampled at 24 equally spaced normalized phases using distinct nearest original frames. No temporal interpolation mixes hidden images into observed inputs. Images are resized bilinearly to $112\times112$, and label-1 cavity masks by nearest-neighbor interpolation. Each target's intensity mean and standard deviation are calculated from its observed frames only. The time axis is normalized cycle phase, not calibrated elapsed seconds. Spatial distances are in resized pixels, not millimeters.

The fixed five-fold split assigns each patient to exactly one test fold. For test folds 0--4, train/validation/test counts are respectively 58/20/20, 58/20/20, 59/19/20, 60/19/19, and 59/20/19. Training and validation patients are disjoint from that fold's test patients. Patients can be sources in other folds, as in ordinary cross-validation. Fold-trained models are not independent source-cohort replicates.

The observed-image budgets are $K=2,3,5,7,10$. Each patient--budget pair has one uniform, five random, and five clustered sampling patterns. Clustered sampling selects a circular interval of length $\max\{K,\lceil0.35T\rceil\}$ on the $T=24$-frame grid, then samples distinct frames within it. Thus the $K=10$ cluster spans at least 10 frames. Patient splits and observed-frame indices are held fixed across training seeds. All methods use the same prespecified observed-frame indices. Hidden target images and all target masks are excluded from neural inference. Target masks are used only for scoring, while source masks supervise training.

\subsection{Learned observation model and conditional inference}
For observed frame $x_i$ at phase $t_i$, a shared encoder produces $\vect{m}_i\in\R^{16}$ and log variances $\vect{v}_i\in[-7,3]^{16}$. It has three stride-two convolution blocks with 24, 48, and 96 channels, GroupNorm and SiLU, followed by spatial average pooling and two linear heads. Let
\[
\phiv(t)=(1,\cos2\pi t,\sin2\pi t,\ldots,\cos6\pi t,\sin6\pi t)^\top.
\]
The coefficient prior is independent across latent dimensions $d$, with learned mean $\vect{w}_{0d}\in\R^7$ and a full positive-definite $7\times7$ covariance $\vect{\Lambda}_d$. A Cholesky parameterization gives positive diagonal entries by softplus plus $10^{-4}$. There is no cross-latent-dimension covariance. A phase head maps the across-shot mean and standard deviation of encoder means to $\delta\in[-0.25,0.25]$ cycles. All design rows use $(t_i+\delta)\bmod1$.

Conditional on these outputs, define $\mat{R}_d=\diag(\exp v_{1d},\ldots,\exp v_{Kd})$. The implemented working posterior is
\begin{align}
\vect{\Sigma}_d&=(\vect{\Lambda}_d^{-1}+\mat{F}^\top\mat{R}_d^{-1}\mat{F})^{-1},\nonumber\\
\widehat{\vect{w}}_d&=\vect{\Sigma}_d(\vect{\Lambda}_d^{-1}\vect{w}_{0d}+\mat{F}^\top\mat{R}_d^{-1}\vect{m}_{:d}).
\label{eq:tedposterior}
\end{align}
The decoder maps the posterior-mean latent state at each phase to mask logits through a linear $96\times4\times4$ representation and transposed-convolution blocks, followed by resizing to the evaluation grid. This is discriminatively trained Gaussian conditioning on encoded observations. It is not a generative likelihood for ultrasound pixels, nor a posterior integrating neural-weight, phase, or hyperparameter uncertainty.

Training minimizes
\begin{equation}
\mathcal L=\mathcal L_{\rm BCE}+\mathcal L_{\rm Dice}
+0.1\mathcal L_{\rm obs}+0.001\mathcal L_{\rm KL}
+0.05\mathcal L_{\rm cycle}+0.01\mathcal L_{\rm temporal}.
\label{eq:tedloss}
\end{equation}
Segmentation losses use full-cycle source masks and predictions at the posterior-mean coefficients. The observation term averages
$\tfrac12\{(m_{id}-\phiv(t_i+\delta)^\top\widehat{\vect{w}}_d)^2\exp(-v_{id})+v_{id}\}$ over shots and latent dimensions. It is a fitted-residual objective, not the marginal likelihood integrating coefficients. The KL is the Gaussian coefficient posterior-to-prior divergence. Cycle loss penalizes the squared probability difference between the first and last sampled frames. These are adjacent cycle endpoints, not identical phases. Temporal loss is the mean absolute second difference of probabilities over consecutive sampled frames.

AdamW uses learning rate $3\times10^{-4}$, weight decay $10^{-5}$, batch size 4, at most 80 epochs, and patience 12. Mixed sparse-frame strategies are sampled during training. Uniform validation observations determine the checkpoint by mean full-cycle Dice. The six original configurations are trained separately for each fold and budget and compared over three training seeds. The design has $6\times5\times5\times3=450$ original-configuration training combinations. The additional noise-scale setting is reported separately from these six configurations. The earlier single-seed diagnostic run used seed 20260825, Python 3.12.7, PyTorch 2.7.1 with CUDA 12.6, and an NVIDIA RTX 4090 D. These environment records refer to that earlier run, not a verified environment inventory for all reruns.

\paragraph{Baseline meanings.}
The weak-prior variant sets $\vect{w}_{0d}=0$ and $\vect{\Lambda}_d=10^4\mat{I}$, while retaining a source-trained encoder, decoder, and phase head. The fixed-coefficient variant sets coefficients to learned prior means but still estimates target phase from observed images. Neither is the same as the photon study's no-source or no-target-photon comparator. The latent-MSE variant replaces only $\mathcal L_{\rm obs}$ by unweighted latent reconstruction MSE. Its inference still uses encoded variances and Eq.~\eqref{eq:tedposterior}. The isotropic-prior variant uses one learned shared variance. The no-phase variant fixes $\delta=0$. The two non-hierarchical coefficient modes omit the KL term.

\paragraph{Observation-noise scale sensitivity.}
The last row of Table~\ref{tab:ted} reports OA-HB-TPM with observation-noise scale 4.0. We label it by its scale rather than as a successfully calibrated model. The aggregate result records a setting and its performance, but does not specify whether the multiplier acts on variance or standard deviation, whether it is applied during training or inference, or which split selected it. We therefore treat this row as a reported sensitivity setting, without assuming a particular modification of Eq.~\eqref{eq:tedposterior} or validation-only selection.

\paragraph{Numerical safeguards.}
The implementation solves weak-prior precision systems in float64 on CUDA/CPU before casting back, because low-$K$ designs are rank deficient and float32 can lose the $10^{-4}$ prior precision. Covariances are symmetrized. Cholesky factorization used in KL/sampling first tries the symmetric covariance and then scale-relative roundoff jitter on failure. These safeguards address numerical positive definiteness, not statistical calibration. Their presence does not explain away the observed undercoverage.

\subsection{Metrics and patient-level inference}
Hidden-frame Dice and HD95 compare thresholded posterior-mean mask probabilities ($p\geq0.5$) with expert masks. HD95 pools both directions of surface distances. A single empty mask receives the resized image diagonal as its penalty, and two empty masks receive zero. Area error uses \emph{soft} predicted areas $\widehat A_t=\sum_x p_t(x)$ and binary reference areas $A_t$ across the whole cycle, including observed phases:
\[
e_A=\frac{\{T^{-1}\sum_t(\widehat A_t-A_t)^2\}^{1/2}}{\max\{T^{-1}\sum_t A_t,1\}}.
\]
It is not an average of pointwise relative errors and not a hidden-frame-only endpoint. Brier scores use hidden-frame pixel probabilities. Posterior-predictive Brier uses the mean of sampled probabilities and is separate from the deterministic decoder-at-mean score.

For each method and training seed, raw replicates are first averaged within patient--budget--pattern cells. The $98\times5\times3$ cells then receive equal weight, followed by equal weighting of the three seed means. Pooling all raw rows would instead over-weight random and clustered patterns fivefold. Repeated seeds do not increase the cohort from 98 to 294 independent patients. Table~\ref{tab:ted} supplies rounded aggregate means. Patient-paired confidence intervals, per-seed dispersion, and per-budget or per-pattern rerun summaries cannot be recovered from those means alone.

From the reported means, original OA improves over weak-prior adaptation by $0.0007$ Dice, $0.0553$ pixels in HD95, and $0.0198$ in area RMSE. HB-TPM with latent-MSE improves over original OA by $0.0058$ Dice, $0.4624$ pixels in HD95, and $0.0371$ in area RMSE. Fixed source coefficients also outperform original OA on each mean reconstruction metric. Noise scale 4.0 lowers OA HD95 by $0.0063$ pixels and area RMSE by $0.0012$, with identical Dice to four decimal places. These differences are arithmetic contrasts of rounded means, not significance estimates.

\begin{figure}[htbp]
\centering
\includegraphics[width=\linewidth]{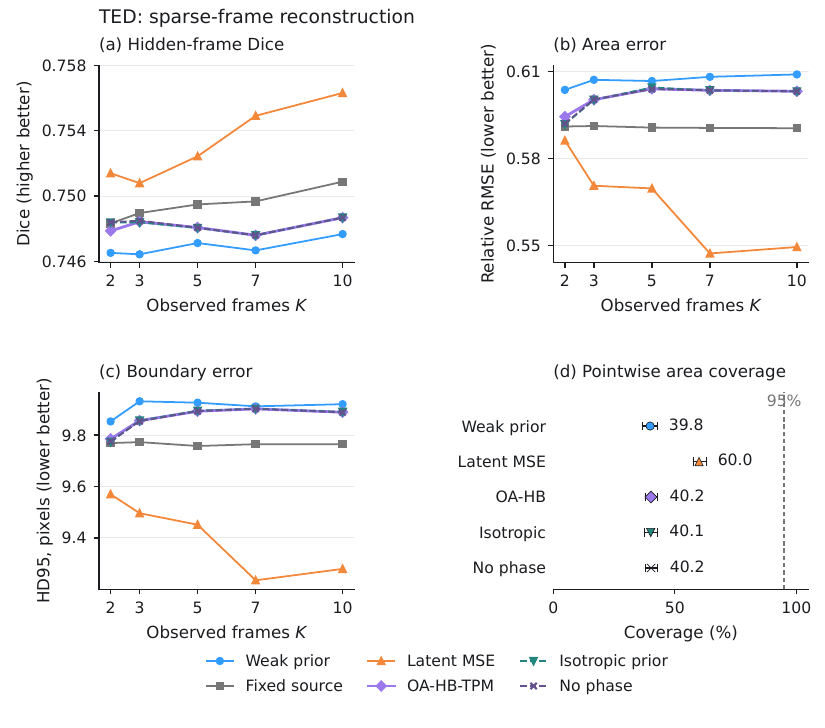}
\caption{\textbf{TED: original single-seed budget diagnostics (seed 20260825).} These panels describe the earlier run, not the three-seed means or noise-scale setting in Table~\ref{tab:ted}. (a--c) Mean hidden-frame Dice, soft-area relative RMSE, and HD95. Replicates are first averaged, with patients and patterns equally weighted. Lines connect separately trained models at discrete budgets. The three OA variants nearly overlap. (d) Pointwise soft-area coverage for the five systems producing intervals. Fixed coefficients produce no intervals. Whiskers use 20000 within-fold patient resamples, conditional on the original fitted models. The dashed line marks nominal 95\% coverage. All 98 patients and patterns are retained.}
\label{fig:teddiagnostics}
\end{figure}

For a paired rerun analysis, seed-averaged cell scores should first be combined within each patient. Resampling whole patients within their fixed test folds then preserves pairing across methods, budgets, and patterns. Variation across training seeds should be reported separately. The earlier single-seed intervals in Figure~\ref{fig:teddiagnostics} are not confidence intervals for Table~\ref{tab:ted}.

\subsection{Uncertainty, rejection, and the contour diagnostic}
Under the original evaluation protocol, 32 Gaussian coefficient draws are decoded to mask probabilities. Soft-area 2.5th and 97.5th percentiles define pointwise intervals. Pixel Bernoulli, neural-weight, and phase variation are not sampled. Coverage averages hidden-frame expert-area inclusion with the same cell weighting as accuracy, not simultaneous cycle coverage. The updated three-seed coverage is $43.6\%$ for weak-prior adaptation, $63.6\%$ for latent-MSE, $39.3\%$ for original OA, $39.1\%$ for isotropic prior, $39.3\%$ without phase alignment, and $42.7\%$ at noise scale 4.0. Fixed coefficients have no sampled intervals. The scale setting improves OA coverage by 3.4 percentage points, while remaining 52.3 points below nominal coverage. Interval widths and stored endpoints are needed to assess the coverage--sharpness trade-off and independently check membership.

\paragraph{Original single-seed auxiliary diagnostics.}
The gate averages variance-standardized latent residuals on observed frames. A threshold at the 95th percentile of uniform-pattern validation-patient scores is used for every target pattern. In the original single-seed run, OA-HB-TPM flags $0.0571$, $0.6731$, and $0.8788$ of uniform, random, and clustered patterns. These rates have not been updated to three-seed estimates. The gate records a flag without fallback or case exclusion. External acquisition, vendor, and trajectory-family shifts are untested.

The retained oracle-contour diagnostic uses manual masks at the same observed phases, fits per-pixel signed-distance fields with periodic Fourier ridge regression, and thresholds them. Its original equal-cell Dice is $0.9375$, HD95 $2.290$ pixels, and area error $0.09264$. Privileged labels and hard rather than soft areas make this neither an image-input competitor nor a formal upper bound. It remains an auxiliary comparison, not a three-seed image-model result.

\subsection{Interpreting the observation-loss and adaptation comparisons}
\label{app:tedinterpretation}
Table~\ref{tab:ted} separates three questions. Original OA improves the aggregate weak-prior comparison. Latent-MSE retains the hierarchical posterior update and gives the strongest mean reconstruction within the tested framework. Fixed coefficients test adaptation against a source template with target-dependent phase and achieve better mean reconstruction than original OA. The isotropic-prior and no-phase rows differ little from original OA. Thus the means favor the latent-MSE configuration without establishing a general benefit for the variance-weighted training loss or a measurable contribution from each OA component.

Two mechanisms are plausible and testable. First, encoded observations can introduce coefficient error when their reliability is overestimated. In Eq.~\eqref{eq:tedposterior}, predicted variances set the weight of each image encoding. Second, the fitted-residual observation loss jointly changes the latent means, variances, prior, and decoder during training, whereas latent-MSE changes that training pressure. These mechanisms are hypotheses: separately trained checkpoints do not isolate either effect. The contour diagnostic additionally measures reconstruction with privileged mask inputs and therefore cannot identify the cause of the image-input gap.

A discriminating test would freeze one source-trained encoder, decoder, phase head, and prior, then compare fixed coefficients, the existing update, and explicitly validation-selected observation variances on exactly the same target frames. The scale-4.0 aggregate shows a small coverage increase, but its selection and execution metadata are needed to connect it to this controlled test. Paired patient-level changes by budget and pattern, per-seed scores, interval widths, and saved predictions would distinguish training variation from adaptation and calibration effects. Any further scale tuning should use validation patients, with test masks reserved for final scoring.

\subsection{Reproducibility and remaining scope}
The three-seed table is transcribed from the reported rerun aggregates. The earlier single-seed records support Figure~\ref{fig:teddiagnostics} and the auxiliary diagnostics, not a reaggregation of the new table. The current aggregate summary does not include per-seed or patient-level rerun records, interval endpoints, exact seed identifiers, or noise-scale selection metadata. Consequently, the manuscript reports means and their arithmetic differences without new bootstrap intervals or claims of seed-wise consistency. Releasing the corresponding run manifests, scores, predictions, and interval bounds will enable those checks. Image-input trajectory baselines, external cohorts, and explicit family shifts remain additional evaluation targets.

\clearpage
\section{Reproducibility and evaluation protocol}
\label{app:reproduce}
The supplement specifies the vector data-generating family, source fitting, baselines, and metrics in Appendices~\ref{app:proofs}--\ref{app:control}. Appendix~\ref{app:photon} details photon simulation and event inference. Appendix~\ref{app:ted} describes cardiac preprocessing, training, evaluation, and resampling.

\paragraph{Data and information boundaries.}
The existing anonymous code supplement includes the vector simulator and results, photon arrival times and fitted coefficients, and the original single-seed cardiac scores with fixed splits and observed-frame indices. Those cardiac records are distinct from the updated three-seed aggregates in Table~\ref{tab:ted}. Photon simulation truth is separated from estimator inputs and used only for evaluation. Cardiac target masks are excluded from inference, while source masks provide training supervision. TED images and annotations must be obtained from the dataset provider under its terms. Raw patient arrays and large checkpoints are not redistributed.

\paragraph{Paired evaluation.}
Vector methods share targets and main-benchmark projection directions. The vector runner uses a sequential random stream across sampling and generation, so exact reproduction requires preserving method order. Changing it can change subsequent targets. The Gaussian-control exception to common projection directions is documented in Appendix~\ref{app:control}. Photon contrasts resample whole paired target records and condition on one learned source prior. TED replicates are averaged within patient--budget--pattern cells, then cells and training seeds receive equal weight. The three-seed table reports aggregate means without newly estimated bootstrap intervals. Patient-level pairing within test folds remains the required unit for subsequent paired resampling. Recorded configurations are not evidence of preregistration.

\paragraph{Software and reproducibility scope.}
Environment requirements and portable execution commands accompany the existing code supplement. Vector and photon computations use Python numerical libraries, and cardiac training uses PyTorch. The vector benchmark has been reproduced. Photon summaries and the original cardiac diagnostics were recomputed from experiment-level records. The updated cardiac table instead uses the reported three-seed aggregates, whose patient-level records and calibration metadata remain to be reconciled with the source package. Its means do not establish independent replication of the full data-to-prediction pipeline. Interval endpoints, prediction-level outputs, and version-linked execution metadata are needed for independent coverage checks.

\clearpage
\section{Fixed-representation image specialization and smoothness diagnostics}
\label{app:imagespecialization}
We relate image-space sampling to the coefficient model. The latent-space views in Appendix~\ref{app:legacyrepresentation} motivate representing structured variation, not identifying an entire image class with a curve. Here a trajectory is a task-specific support approximation within a representation that may have several varying factors. The following specializations and diagnostics do not constitute additional MNIST or CelebA experiments.

\paragraph{Fixed representations.}
Fix an encoder $e_\theta:\mathcal X\to\R^D$ and decoder $d_\psi:\R^D\to\mathcal X$ before target adaptation. Set $\vect{z}_i=e_\theta(x_i)$ for the target images, apply Eq.~\eqref{eq:objective} to these features, then decode:
\begin{equation}
\widetilde x=d_\psi\!\left(\widehat{\mat{W}}^{\top}\phiv(U)+\vect{\eta}(U)\right),
\qquad U\sim\operatorname{Unif}[0,1],
\label{eq:decodedimages}
\end{equation}
where $\vect{\eta}$ is the perturbation in Eq.~\eqref{eq:sampling}. For identical encoded inputs, basis, prior, and assignments, coefficient inference is unchanged. Decoding is subsequent processing. Identity maps recover the vector setting. Learning the prior still requires comparable observed source progression (Section~\ref{sec:sourcefit}), not an arbitrary ordering of static images. The Gaussian likelihood describes encoded vectors, not pixels. Decoder fidelity and class preservation require separate evaluation. Unlike Section~\ref{sec:imageextension}, the representation is fixed rather than jointly trained.

\paragraph{A non-hierarchical, fixed-prior reduction.}
Instead of learning hyperparameters from sources, choose $\vect{w}_{0d}=\vect{0}$ and $\vect{\Lambda}_d^{-1}=\alpha\mat{I}_p$ for all $d$, with a fixed $\alpha>0$. Conditional on $\vect{t}$, Eq.~\eqref{eq:objective} becomes ridge curve fitting:
\begin{equation}
\mathcal J_\alpha(\mat{W},\vect{t})=
\frac{\|\mat{Z}-\mat{F}_{\vect{t}}\mat{W}\|_F^2}{2\sigma^2}
+\frac{\alpha}{2}\|\mat{W}\|_F^2,
\qquad
\widehat{\mat{W}}_\alpha=
(\mat{F}_{\vect{t}}^\top\mat{F}_{\vect{t}}+\sigma^2\alpha\mat{I}_p)^{-1}
\mat{F}_{\vect{t}}^\top\mat{Z}.
\label{eq:fixedprior}
\end{equation}
Differentiating $\mathcal J_\alpha$ gives the conditional MAP, not equivalence of initialization or coordinate solvers. This is not hierarchical transfer. As $\alpha\downarrow0$, the displayed zero-centered ridge estimate tends to $\mat{F}_{\vect{t}}^\dagger\mat{Z}$, with $\dagger$ denoting the Moore--Penrose inverse. With full column rank this is ordinary least squares. Otherwise it is the minimum-norm solution, but the flat-prior posterior is improper. For a general prior mean, Appendix~\ref{app:flatlimit} gives $\widehat{\vect{w}}_\alpha\to\mat{F}^\dagger\vect{z}+(\mat{I}_p-\mat{F}^\dagger\mat{F})\vect{w}_0$. Thus weakening a prior retains its mean in null directions: for $\mat{F}=(1,0)$ and $\vect{w}_0=(a,b)^\top$, the second coefficient stays $b$, while its variance $1/\alpha$ diverges. The finite weak-prior ablation is not a no-transfer estimator.

\paragraph{What is not an exact reduction.}
PCA-ordered interpolation, curvature-penalized splines, and joint classification/representation objectives are related trajectory methods, not automatically this MAP estimator. Exact reduction requires matching the basis, objective, coordinate assignments, and sampler. Decoder composition alone does not make existing image panels evidence for hierarchical transfer.

\paragraph{Smoothness is a diagnostic, not an estimator.}
For an ordered grid $u_1<\cdots<u_M$ with $M\geq2$, define noise-free decoded points $x_m=d_\psi(\widehat{\mat{W}}^\top\phiv(u_m))$ and their mean adjacent displacement
\begin{equation}
S_M=\frac{1}{M-1}\sum_{m=1}^{M-1}\|x_{m+1}-x_m\|_2.
\label{eq:imagesmoothness}
\end{equation}
For digit or face trajectories, comparisons must fix $M$, grid placement, endpoints, and pixel normalization or evaluation representation. If the decoded path has finite length $L_{\rm path}$, the definition of arc length gives $S_M\leq L_{\rm path}/(M-1)$: finer grids alone drive this metric to zero. A collapsed output also has $S_M=0$. Lower displacement alone establishes neither realism, diversity, class consistency, nor downstream utility. Permuting the same decoded points isolates ordering, not a separately fitted generator's quality. Evaluate sampling noise separately from backbone smoothness.

Empirical evaluation requires source/target splits, checkpoints, generated outputs, and per-run scores. Report smoothness alongside reconstruction, diversity/class consistency and, when claimed, held-out downstream performance. Mathematical inclusion is not empirical validation. Appendix~\ref{app:legacy} retains earlier digit/face illustrations and a target-only spiral pilot, with their distinct evidential roles stated explicitly.

\clearpage
\section{Earlier non-hierarchical TPM studies}
\label{app:legacy}
Earlier non-hierarchical TPM studies examine representation geometry, interpolation smoothness, and classification. We retain these image studies alongside a reproduced spiral pilot. The image figures and reported summaries remain available, but their underlying splits, checkpoints, and run-level outputs are incomplete. They provide context for the trajectory approach rather than tests of hierarchical transfer. Appendix~\ref{app:imagespecialization} distinguishes exact reductions from related trajectory estimators.

\subsection{Digit and face interpolation}
\label{app:legacyimages}
\begin{figure}[htbp]
\centering
\includegraphics[width=\linewidth]{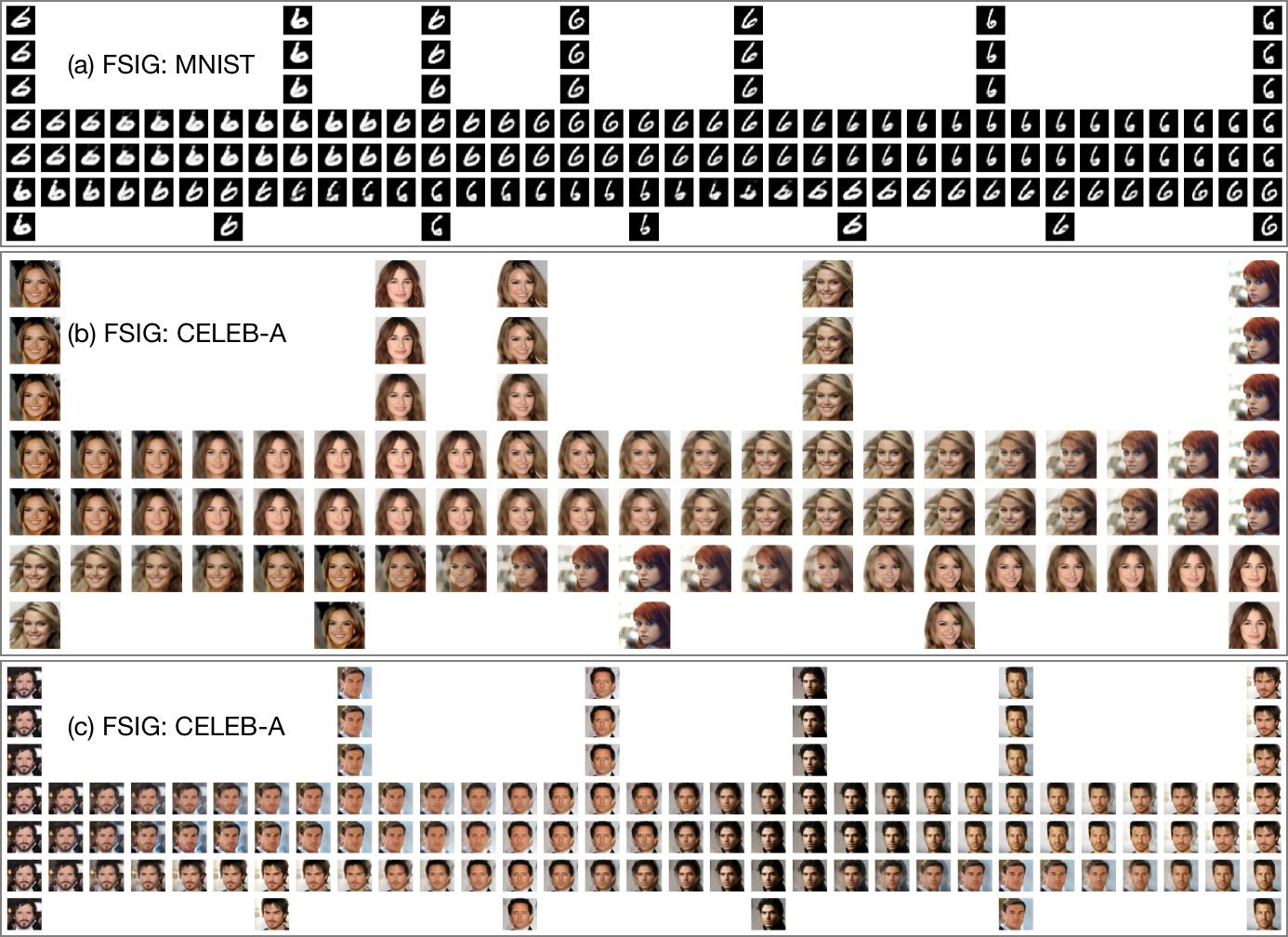}
\caption{Earlier digit and face interpolation examples. Within each of the three blocks, the seven rows retain the original descriptions: (1) image anchors, (2) autoencoder reconstructions, (3) reconstructions of curve-projected codes, (4) linear interpolation in the estimated order, (5) spline interpolation in that order, (6) linear interpolation in a shuffled order, and (7) anchors in the shuffled order. Anchor counts differ across blocks. This is a qualitative illustration from the earlier TPM workflow, not an HB-TPM experiment. Dataset attributions and row descriptions follow the original record. Sample-level identities, splits, checkpoints, and intermediate outputs are unavailable for independent regeneration.}
\label{fig:legacyimages}
\end{figure}

Figure~\ref{fig:legacyimages} preserves digit and face examples previously attributed to MNIST \citep{lecun1998gradient} and CelebA \citep{liu2015faceattributes}. Its purpose is to make image-space trajectory sampling tangible: intermediate decoded images connect sparse anchors, and changing their order changes the path. These selected examples motivate the smoothness diagnostic in Eq.~\eqref{eq:imagesmoothness}, but do not measure its population-level behavior.

\paragraph{Interpretation.}
The montage illustrates appearance transitions, not physical temporal progression. Static face examples need not constitute an observed source trajectory. Smooth-looking interpolants alone do not establish identity preservation, distributional fidelity, diversity, or classification gains. Sections~\ref{app:legacysmoothness}--\ref{app:legacyclassification} restore the historical smoothness and classification records separately, with their verification limits. No numerical performance claim is inferred from this montage. The current tests of hierarchical transfer remain those in Section~\ref{sec:experiments}.

\FloatBarrier
\clearpage
\subsection{Target-only spiral classification pilot}
\label{app:legacypilot}
Table~\ref{tab:legacypilot} retains all seven methods and both budgets from a single-seed target-only prototype. A rerun with Python 3.12.4 and NumPy 2.5.2 reproduces all 14 accuracies and recorded curve diagnostics exactly. It is not an independent repeat.

\begin{table}[htbp]
\centering
\caption{\textbf{Synthetic three-class spirals: non-hierarchical pilot.} Test accuracy (\%) over 2,100 points, one seed per cell. Bold marks the highest accuracy. There are no across-seed error bars. This is a target-only prototype study, not a matched HB-TPM ablation.}
\label{tab:legacypilot}
\begin{tabular}{lrr}
\toprule
Method & $K=3$ & $K=5$ \\
\midrule
No augmentation & 60.29 & 56.86 \\
Gaussian augmentation & \textbf{63.10} & 57.43 \\
Pairwise interpolation & 58.90 & 56.52 \\
Random-order curve & 49.10 & 56.90 \\
PCA line & 56.71 & 57.76 \\
Offline TPM prototype & 56.14 & \textbf{58.57} \\
Joint TPM prototype (NumPy) & 58.52 & 55.90 \\
\bottomrule
\end{tabular}
\end{table}

\begin{figure}[htbp]
\centering
\includegraphics[width=\linewidth]{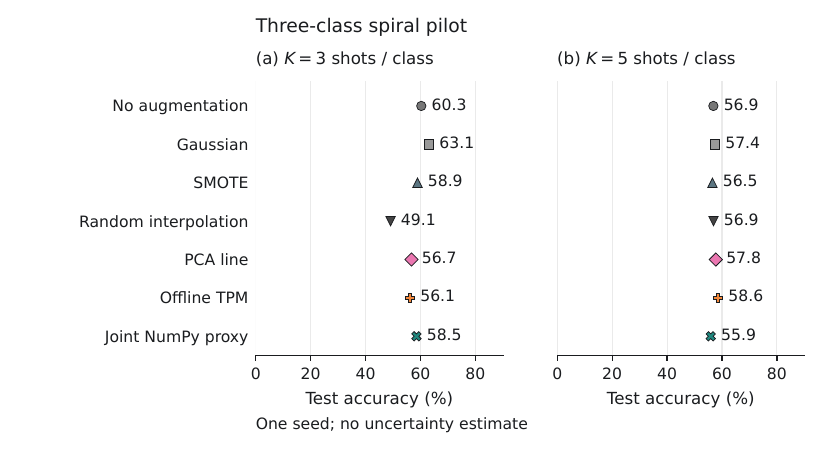}
\caption{\textbf{Three-class spirals: all methods in the rerun pilot.} Test accuracy for $K=3$ and $K=5$ labeled shots per class, using one seed and 700 test points per class. Higher is better. Points reproduce Table~\ref{tab:legacypilot}, with no inferred error bars. The shot sets are not nested, and the joint proxy has different training and preprocessing. These target-only prototypes are not hierarchical-transfer baselines. The common axis starts at zero to show the scale of the differences.}
\label{fig:spiralpilotcomparison}
\end{figure}

\paragraph{Protocol.}
Three spiral arms supply 360 training-pool and 700 independent test points per class. Each arm uses 900 positions, radius $0.06$--$1.25$, a four-radian span, and coordinate scale $1.12$, with normal/tangential Gaussian noise scales $0.075/0.035$. Methods train on shared $K=3$ or $5$ labeled shots per class, with separately drawn, non-nested budgets. Augmenters add 60 samples per class. Curve samplers use five B\'ezier controls and normal noise scale $0.07$.

\paragraph{Methods and limitations.}
Offline TPM initializes a curve from PCA-ordered shots without source transfer or a coefficient MAP update. The joint NumPy proxy alternates softmax fitting with finite-difference curve updates using classification, projection, and curvature terms. Other methods use 250 classifier updates and standardize augmented data. The proxy uses 80 updates and standardizes shots alone. Thus compute and preprocessing are not matched. All classifiers are linear, restrictive for curved classes. These prototypes are not exact reductions of HB-TPM (Appendix~\ref{app:imagespecialization}).

Offline TPM exceeds no augmentation only at $K=5$. The joint proxy does not at either budget. One seed gives no between-run uncertainty. These mixed results motivate matched multi-seed tests, not a general augmentation or hierarchical-transfer advantage.

\begin{figure}[htbp]
\centering
\includegraphics[width=0.96\linewidth]{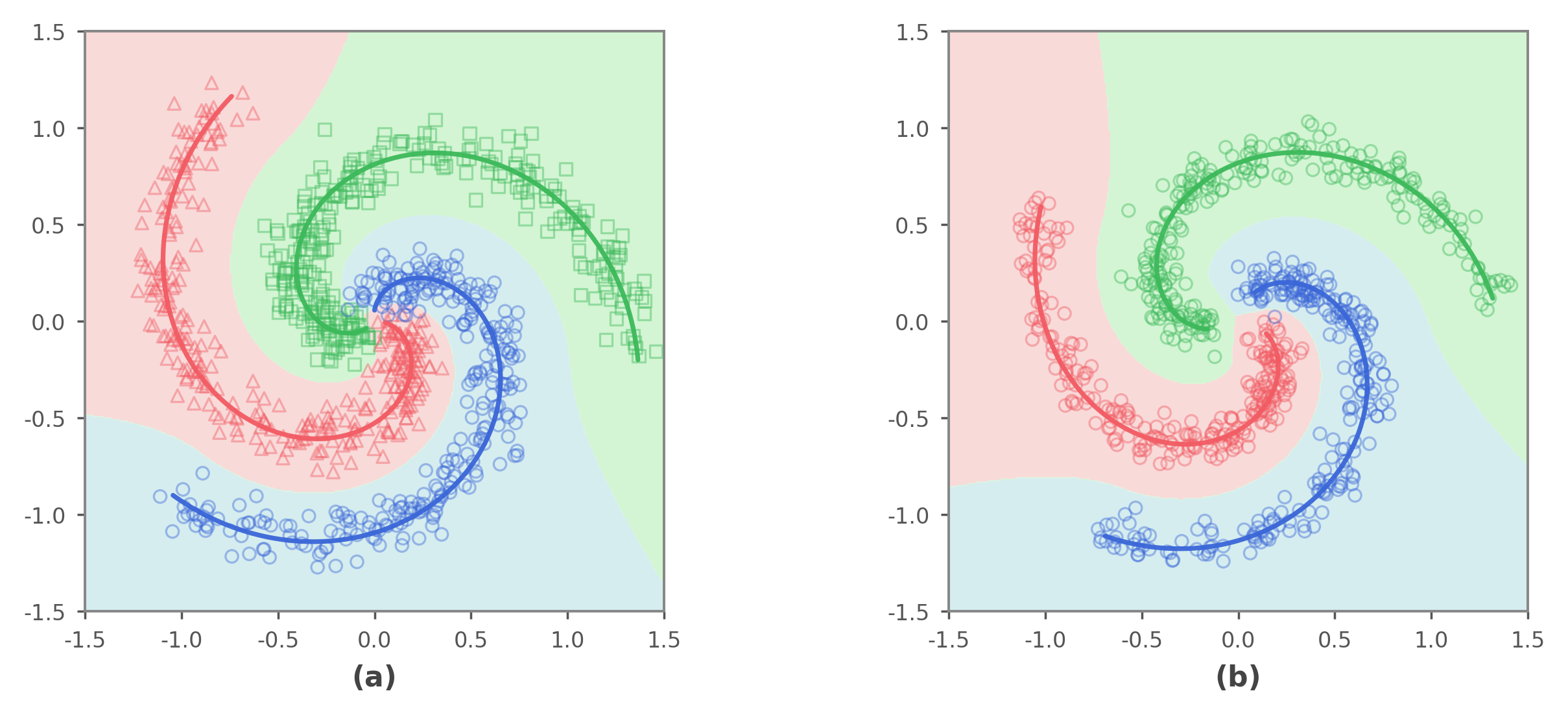}
\caption{Prescribed spiral geometry (schematic), not fitted classifiers. (a) Three fixed curves with sampled perturbations. (b) Fixed truncations, rescalings, and shifts, followed by resampling. Backgrounds use nearest-curve distances plus a fixed analytic bias. Neither panel depicts the fitted pilot models or provides accuracy evidence for Table~\ref{tab:legacypilot}.}
\label{fig:legacyspiral}
\end{figure}
\FloatBarrier
\clearpage
\subsection{Latent geometry as motivation for trajectory sampling}
\label{app:legacyrepresentation}
\begin{figure}[htbp]
\centering
{\small\makebox[0.31\linewidth]{Raw pixels}\hfill\makebox[0.31\linewidth]{Standard AE}\hfill\makebox[0.31\linewidth]{Regularized AE}\par}
\smallskip
\includegraphics[width=\linewidth]{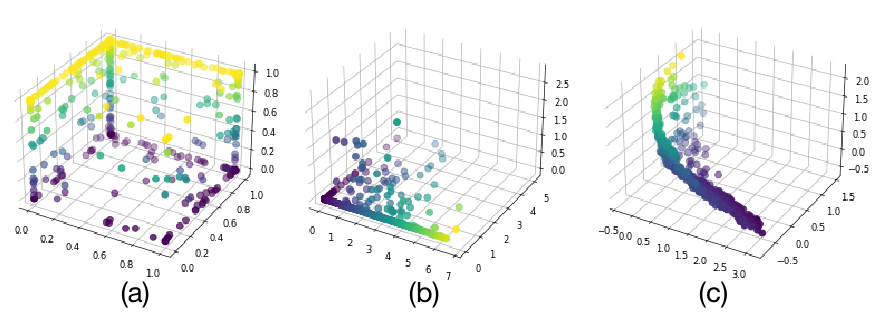}
\caption{Earlier MNIST representation views (original descriptions). The curved pattern in (c) motivates a structured support approximation, not one-dimensional class support. Axes and scales differ. Embeddings, sample indices, and projection metadata are unavailable. These unchanged archival panels provide geometric intuition, not evidence of hierarchical-transfer gains.}
\label{fig:legacyrepresentation}
\end{figure}

Figure~\ref{fig:legacyrepresentation}(c) displays a curved, concentrated pattern, qualitatively motivating geometry-aware sampling. Regularization encourages this pattern, so it is not independent evidence for a natural manifold. Table~\ref{tab:legacydimension} reports dimensions 3--5, not one: a curve approximates task-specific progression rather than all within-class variation. Approximation accuracy and source-prior utility require separate tests.

The recorded autoencoder used 128-dimensional codes, five convolutional layers per encoder/decoder plus a fully connected encoder layer, Adam ($10^{-3}$), and batch size 128. Neighborhood settings, regularization strength, checkpoints, and splits are unavailable. Dimensions have not been re-estimated.

\begin{table}[htbp]
\centering
\caption{\textbf{MNIST: intrinsic dimensions reported in the earlier study.} Entries are transcribed from the prior manuscript, not re-estimated from retained embeddings. No uncertainty estimates or exact neighborhood settings are available. Lower dimension is not a performance ranking. TPM feature refers to the earlier representation regularizer, not hierarchical trajectory transfer.}
\label{tab:legacydimension}
\begin{tabular}{lrrrrrrrrrr}
\toprule
Digit & 0 & 1 & 2 & 3 & 4 & 5 & 6 & 7 & 8 & 9 \\
\midrule
Raw input & 11 & 9 & 12 & 12 & 12 & 13 & 11 & 10 & 14 & 11 \\
AE feature & 10 & 8 & 10 & 11 & 10 & 11 & 9 & 9 & 11 & 10 \\
TPM feature & 4 & 3 & 5 & 5 & 5 & 5 & 4 & 4 & 5 & 4 \\
\bottomrule
\end{tabular}
\end{table}

\begin{figure}[htbp]
\centering
\includegraphics[width=\linewidth]{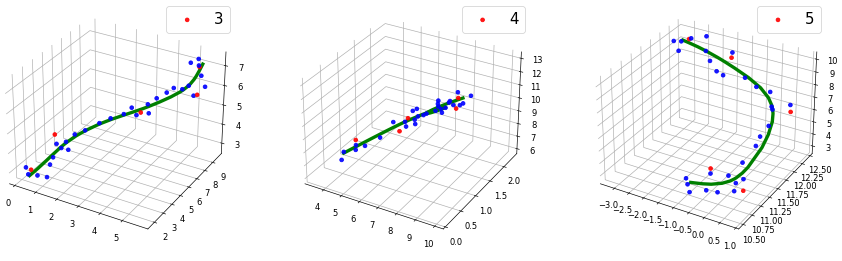}
\caption{Earlier target-only MNIST curves for digits 3, 4, and 5. Per the original description, red markers denote five shots, green curves reconstructions, and blue markers generated codes. The displays illustrate sparse-observation organization, not hierarchical transfer. Exact embeddings and fitted coefficients are unavailable.}
\label{fig:legacymnistcurves}
\end{figure}
\FloatBarrier
\clearpage

\subsection{Earlier digit and face smoothness results}
\label{app:legacysmoothness}
\begin{figure}[htbp]
\centering
\textnormal{(a) MNIST: original recorded smoothness panel}\par\smallskip
\includegraphics[viewport=0 0 828 420,clip,width=0.90\linewidth]{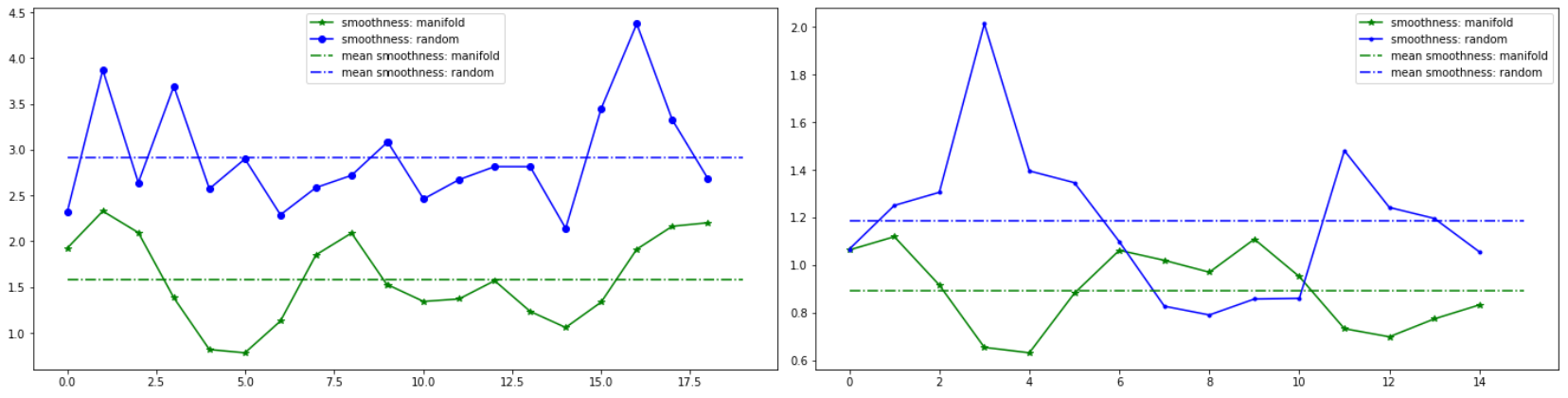}
\par\medskip
\textnormal{(b) CelebA: original recorded smoothness panel}\par\smallskip
\includegraphics[viewport=828 0 1656 420,clip,width=0.90\linewidth]{figures/mnist_celeb_smoothness.png}
\caption{Smoothness summaries from the earlier MNIST and face studies. The two original panels are arranged vertically without changing their curves. Green stars denote manifold ordering and blue circles denote random ordering. Horizontal dash-dot lines are the means shown in the original panels, not uncertainty intervals. Lower values describe smaller reported adjacent changes, not higher image quality by themselves. Underlying image sequences and run-level metric records are unavailable for independent rescoring.}
\label{fig:legacysmoothness}
\end{figure}

Figure~\ref{fig:legacysmoothness} restores the original smoothness panels without later-added shaded bands. The earlier study defined smoothness as mean adjacent decoded-image $L^2$ displacement (Eq.~\eqref{eq:imagesmoothness}) and compared trajectory with shuffled ordering. Displayed means favor trajectory ordering, but not every individual setting does.

Evaluation-unit definitions, pixel normalization, sequence lengths, and run counts are incomplete. The traces have not been rescored or digitized. We do not compare absolute scales across datasets or attach confidence intervals. Grid density and collapse can also lower displacement without improving quality (Appendix~\ref{app:imagespecialization}).

\FloatBarrier
\clearpage

\subsection{Recorded few-shot classification summaries}
\label{app:legacyclassification}
\begin{figure}[htbp]
\centering
\includegraphics[width=\linewidth]{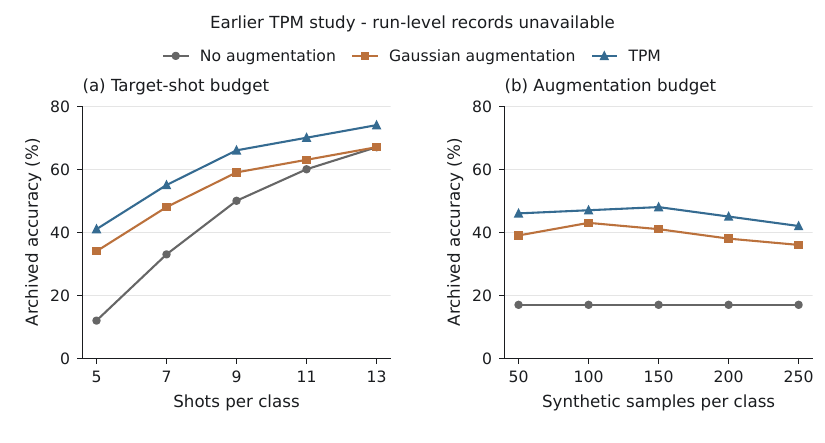}
\caption{MNIST accuracy summaries from the earlier plotting record. The redraw preserves its point values and method labels, without adding uncertainty bands. The horizontal no-augmentation reference in (b) is the arithmetic mean, 17.0\%, of five values in that record, whose relation to individual experimental runs is undocumented. The training/evaluation split and repeated-run outputs are unavailable. This figure reconstructs the reported summary, not an independently replicated experiment.}
\label{fig:legacyclassification}
\end{figure}

Figure~\ref{fig:legacyclassification} restores the MNIST classification values preserved in the earlier plotting record. The shot-budget panel contains $K=5,7,9,11,13$. The augmentation-budget panel contains 50--250 synthetic samples per class. We preserve all recorded point values, omit the old unsupported uncertainty bands, and retain the old method name TPM rather than relabeling it HB-TPM.

The recorded TPM curve lies above the Gaussian curve at all displayed shot budgets. The recorded augmentation curves peak at 100--150 samples. These are descriptions of the archived series, not independently established improvements. The evaluation split, classifier configuration, shot-selection seeds, repeated-run scores, and the fixed shot count for the augmentation sweep have not been recovered. In particular, the two panels cannot be assumed to share a common no-augmentation baseline or matched experiment. We therefore do not pool these values with the current results, infer significance, or claim a downstream benefit for HB-TPM from them.

\paragraph{Earlier seven-shot spiral report.}
The previous manuscript also reported two directional classification accuracies after augmenting seven shots to 300 samples per class. Table~\ref{tab:legacyspiralreported} preserves that report separately from the reproducible $K=3,5$ pilot. The corresponding classifier, prediction records, test counts, and uncertainty estimates have not been recovered. Neither the pilot nor the prescribed schematic in Figure~\ref{fig:legacyspiral} verifies these two numbers.

\begin{table}[htbp]
\centering
\caption{\textbf{Synthetic spirals: accuracies reported in the earlier study.} Values are transcribed from its numerical report. Training and evaluation directions differ, so the rows do not form a method ranking. Run-level records are unavailable, and these values are distinct from the reproduced pilot in Table~\ref{tab:legacypilot}.}
\label{tab:legacyspiralreported}
\begin{tabular}{llr}
\toprule
Classifier training data & Evaluation data & Reported accuracy \\
\midrule
Real spiral data & Generated samples & 97.89\% \\
Generated samples & Real spiral data & 89.4\% \\
\bottomrule
\end{tabular}
\end{table}
\FloatBarrier
\clearpage

\subsection{Earlier spiral shot-coverage visualization}
\label{app:legacyspiralpanels}
\begin{figure}[htbp]
\centering
\includegraphics[viewport=0 0 916 324,clip,width=\linewidth]{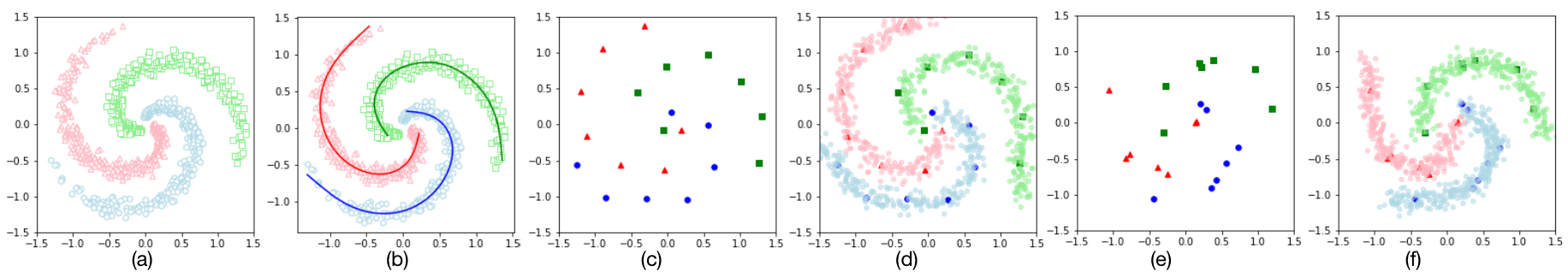}
\par\medskip
\includegraphics[viewport=916 0 1832 324,clip,width=\linewidth]{figures/spiral_manifold_generation_v0.2.png}
\caption{Earlier spiral generation example, retained from the original six-panel image and arranged as two rows for readability. According to its original description, (a) shows reference samples, (b) their class curves, (c) seven representative shots per class, (d) generated samples from the corresponding reconstruction, (e) seven randomly selected shots, and (f) the resulting reconstruction and samples. Colors identify the three classes. These historical panel descriptions have not been checked against retained coordinates or fitted models. The figure is qualitative and supplies no evidence for the numerical accuracies in Table~\ref{tab:legacyspiralreported}.}
\label{fig:legacyspiralpanels}
\end{figure}

The original six-panel spiral example in Figure~\ref{fig:legacyspiralpanels} contrasts representative anchors with a randomly selected set. It makes the coverage issue concrete: a curve fitted within a covered segment and a curve extrapolated beyond the available shots solve different problems. This motivates source-informed recovery, but the panels themselves are from the earlier target-only pipeline, not from HB-TPM or the rerun pilot.

\paragraph{How the earlier studies relate to the present method.}
The representation and interpolation examples motivate learning a useful encoded geometry. The smoothness records motivate checking adjacent variation, and the classification summaries motivate testing downstream utility at several augmentation budgets. The present contribution adds a learned distribution over trajectory coefficients and target adaptation under an explicit observation model. Exact fixed-prior reductions are given in Appendix~\ref{app:imagespecialization}. The earlier objectives and figures are related precursors, not retrospective tests of this hierarchy.

\paragraph{Evidence boundary.}
The supplement retains the original raster assets, an anonymized excerpt of the earlier numerical report, and the classification plotting values and extraction code. These allow exact figure/summary reconstruction where stated, not independent replication of the earlier image experiments. Only the spiral pilot in Table~\ref{tab:legacypilot} was re-executed. The archival records do not enter the abstract's effect sizes or the main comparisons.
\FloatBarrier

\clearpage
\section{Historical context and related comparison families}
\label{app:literature}
The trajectory-transfer problem sits at the intersection of geometric representation, population-level function estimation, and conditional generation. These connections help distinguish the statistical ingredients inherited by HB-TPM from the particular combination evaluated here.

\paragraph{From low-dimensional representations to generative geometry.}
Principal curves describe nonlinear population structure through self-consistency \citep{hastie1989principal}. Generative topographic mapping instead places a probabilistic latent-variable model around a nonlinear map \citep{bishop1998gtm}. Isomap, locally linear embedding, and Laplacian eigenmaps recover low-dimensional representations using geodesic distances, local reconstruction, or graph geometry \citep{tenenbaum2000global,roweis2000nonlinear,belkin2003laplacian}. These are distinct objectives: an embedding need not supply an observation model, a density along a curve, or a transfer rule for a new task. HB-TPM prespecifies a smooth basis and estimates a distribution of task coefficients. Its use of ``principal-manifold'' describes the structured sampling construction, not satisfaction of the classical principal-curve equations. The earlier MNIST displays motivate geometric modeling but do not identify a natural one-dimensional support or validate the learned hierarchy.

\paragraph{From empirical Bayes to sparse functional recovery.}
Empirical-Bayes shrinkage borrows information to stabilize weakly observed parameters \citep{efron1973stein}. Sparse functional-data analysis brings this principle to individual curves, using population means and covariances to reconstruct partially observed trajectories \citep{yao2005functional,wang2016functional}. With fixed target coordinates and Gaussian errors, our finite-basis conditional update is a familiar instance of this idea. The additional vector problem is that the target coordinates are unobserved, making adaptation jointly nonconvex. A comparison with PACE or a multi-task GP should therefore distinguish an observed-coordinate benchmark from a latent-coordinate one. Giving a competitor true target coordinates would answer a different question, while withholding source order from it would confound modeling with information access. Shrinkage can reduce posterior uncertainty under the assumed hierarchy without uniformly reducing frequentist error for every fixed target.

\paragraph{From neural processes to joint function generation.}
Conditional, latent, attentive, and convolutional neural processes learn task-conditioned prediction rules \citep{garnelo2018conditional,garnelo2018neural,kim2019attentive,gordon2020convolutional}. Transformer neural processes and autoregressive CNP deployment address richer predictive dependencies \citep{nguyen2022transformer,bruinsma2023autoregressive}. Neural diffusion processes learn function distributions through finite marginals, with geometric extensions incorporating symmetry priors \citep{dutordoir2023neural,mathieu2023geometric}. Functional diffusion and functional flow matching extend generative modeling to functions on continuous domains \citep{zhang2024functional,kerrigan2024functional}. Score-based neural processes and efficient autoregressive transformer inference further develop conditional function prediction \citep{dou2025score,hassan2026efficient}. These are particularly relevant future comparators because the output is a collection of related predictions, not isolated marginal means. HB-TPM offers an explicit, low-dimensional coefficient update but a more restrictive family. The present experiments do not establish superiority over these models.

\paragraph{Observation models and inverse problems.}
Diffusion posterior sampling combines a learned generative prior with noisy measurement information, including nonlinear inverse problems \citep{chung2023diffusion}. The common principle is to distinguish the latent object from its observation process. Our photon extension uses an event quasi-likelihood and a finite coefficient prior, rather than diffusion-based posterior sampling. Our cardiac extension conditions on neural encodings rather than modeling raw ultrasound intensities. Consequently neither model inherits exact posterior calibration from the general principle of combining a prior and a likelihood. The photon and TED coverage diagnostics test this separately from point-prediction error.

\paragraph{What a broader benchmark must match.}
A fair extension should fix patient or task splits, source progression access, target budgets, preprocessing, and hyperparameter-selection data. For vector data, compare entire generated distributions as well as curve proximity. For photons, keep simulator truth out of estimation and evaluate the same event records. For cardiac data, distinguish hidden-frame reconstruction from segmentation of available images and compare uncertainty at the patient level. Recent few-shot image-generation systems such as DataDream, HypDAE, geodesic-surface transfer, and Uni-DAD address related data scarcity through different architectures, training signals, and uses of pretraining \citep{kim2024datadream,li2025hypdae,han2026few,bahram2026unidad}. In particular, geodesic-surface transfer constructs a pseudo-source domain without requiring a pretrained generative model. These differences motivate information-matched comparisons.

\paragraph{Nearest comparator protocol.}
For fixed coordinates, the coefficient model induces a GP with mean $m_d(t)=\phiv(t)^\top\vect{w}_{0d}$ and kernel $k_d(t,t')=\phiv(t)^\top\vect{\Lambda}_d\phiv(t')$. A GP using this kernel is an equivalence check, not an independent competing model. A substantive GP comparison would learn a different covariance family from the same source patients, such as a periodic kernel for cardiac cycles, with observation noise selected on validation patients. PACE would estimate the source mean and covariance, choose the functional rank on validation patients, and condition functional principal-component scores on sparse target encodings \citep{yao2005functional}. A discretized dense-grid FPCA implementation should be labeled accordingly rather than presented as the full sparse-data PACE estimator.

The cardiac setting provides observed times and a natural common interface: hold the image representation and decoder fixed, and compare trajectory inference from identical encoded frames. A conditional or attentive neural process can learn from source context--query splits under the same patient partition \citep{garnelo2018conditional,kim2019attentive}. Marginal-only CNP predictions require an explicit choice of joint sampling model before comparing full-cycle uncertainty. Separately retrained end-to-end systems should be reported as a second comparison. In the unordered-vector setting, all methods must infer target coordinates, or all must receive coordinates in a separately labeled observed-coordinate diagnostic. These comparisons are proposed experiments and have not been run in the reported benchmark.

\FloatBarrier

\end{document}